\documentclass[11pt]{article}

\usepackage[preprint]{acl}

\usepackage{times}
\usepackage{latexsym}

\usepackage[T1]{fontenc}

\usepackage[utf8]{inputenc}

\usepackage{microtype}

\usepackage{inconsolata}

\usepackage{graphicx}

\usepackage{amsmath,amssymb,mathtools,bm,empheq,amsthm}

\theoremstyle{plain}

\theoremstyle{definition}

\theoremstyle{remark}

\def\bal#1\eal{\begin{align}#1\end{align}} 
\def\suml{\sum\limits}

\newcommand{\pr}[1]{\left(#1\right)} 
\newcommand{\cbr}[1]{\left\{#1\right\}} 
\DeclareMathOperator*{\argmin}{arg\,min} 
\DeclareMathOperator*{\argmax}{arg\,max} 

\DeclareMathOperator*{\rank}{rank}
\def\transp{\mathsf{T}} 
\def\m{\mathbf}
\def\mc{\mathcal}
\def\R{\mathbb{R}}

\def\ast{*}

\newcommand{\grad}[2]{\ensuremath{\nabla_{#2}#1}} 
\newcommand{\norm}[2]{\ensuremath{\left\|#1\right\|_{#2}}}

\newcommand {\bbmtx}{\begin{bmatrix}} 
\newcommand {\ebmtx}{\end{bmatrix}} 

\DeclareMathOperator*{\trace}{tr} 

\usepackage{multirow}
\usepackage{hyperref}
\usepackage{url}
\usepackage{caption}
\usepackage{subcaption}

\definecolor{teal}{RGB}{102, 194, 165}
\definecolor{coral}{RGB}{252, 141, 98}
\definecolor{lavender}{RGB}{141, 160, 203}

\newcommand{\solidline}[1]{\textcolor{#1}{\rule[0.5ex]{1.5em}{0.8pt}}}
\newcommand{\dashedline}[1]{%
  \textcolor{#1}{%
    \texttt{-}\,\texttt{-}\,\texttt{-}%
  }%
}
\usepackage{pifont}
\definecolor{amaranth}{rgb}{0.9, 0.17, 0.31}
\definecolor{americanrose}{rgb}{1.0, 0.01, 0.24}
\definecolor{applegreen}{rgb}{0.55, 0.71, 0.0}
\definecolor{asparagus}{rgb}{0.53, 0.66, 0.42}
\definecolor{53eL}{rgb}{0.0, 0.45, 0.85}
\definecolor{X2Ep}{rgb}{0.55, 0.25, 0.75}
\definecolor{xicA}{rgb}{0.95, 0.5, 0.1}
\newcommand{\cmark}{\textcolor{applegreen}{\scalebox{0.8}{\ding{51}}}}
\newcommand{\xmark}{\textcolor{amaranth}{\scalebox{0.8}{\ding{55}}}}
\usepackage{arydshln}

\usepackage{algpseudocode}
\usepackage{enumitem}
\usepackage{arydshln}
\usepackage{nicefrac}
\usepackage{algorithm2e}
\usepackage{fancyhdr}
\usepackage{fancyhdr}

\renewcommand{\headrulewidth}{0pt}
\renewcommand{\footrulewidth}{0pt}

\fancypagestyle{firstpage}{
    \fancyhf{}
    \fancyfoot[L]{%
        \parbox[b]{0.70\textwidth}{%
            \raggedright\scriptsize
            Published in ACL \textit{Findings of EMNLP 2026},
            Budapest, Hungary.
        }%
    }
    \fancyfoot[C]{\thepage}
    \renewcommand{\headrulewidth}{0pt}
    \renewcommand{\footrulewidth}{0pt}
}

\SetAlgoLined
\SetKwInOut{Input}{Input}
\SetKwInOut{Output}{Output}
\SetKwComment{Comment}{\% }{}   

\title{\hspace{1ex}\includegraphics[width=.45cm]{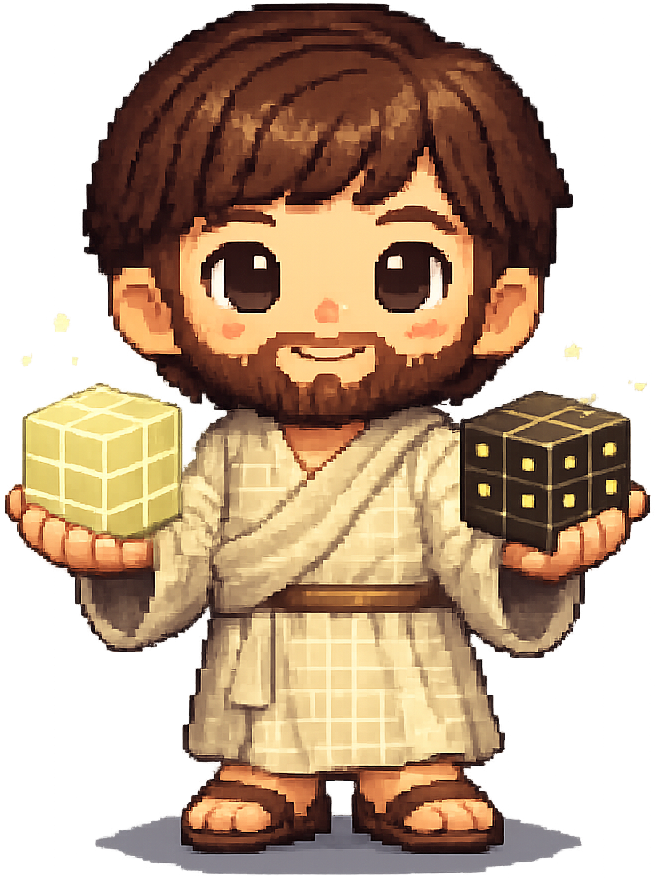} CoSpaDi: \underline{Co}mpressing LLMs via Calibration-Guided \underline{Spa}rse \underline{Di}ctionary Learning}

\author{
 \textbf{Denis Makhov\textsuperscript{*1}},
 \textbf{Dmitriy Shopkhoev\textsuperscript{*1,2}},
 \textbf{Magauiya Zhussip\textsuperscript{1,2}}, \\
 \textbf{Ammar Ali\textsuperscript{1,2}},
 \textbf{Stamatios Lefkimmiatis\textsuperscript{1,2}}
\\
\textsuperscript{1}Fundamental Research Center MWS AI,
 \textsuperscript{2}ITMO
\\
\textit{\url{https://github.com/mts-ai/CoSpaDi}}
}

\begin{document}
\maketitle
$\def\thefootnote{*}\footnotetext{These authors contributed equally to this work\\{\centering\textbf{Correspondence: }\href{mailto:makhovds@gmail.com}{makhovds@gmail.com}}}$
\thispagestyle{firstpage}
\begin{abstract}
Post-training LLM compression often relies on low-rank approximations, which force all columns of a projection matrix to share a single low-dimensional subspace. We propose CoSpaDi, a training-free compression framework that replaces this single-subspace assumption with a union-of-subspaces model via sparse dictionary learning. CoSpaDi factorizes each weight matrix into a dense dictionary and column-sparse coefficients, allowing different columns to select different subsets of dictionary atoms at the same storage budget. To preserve model behavior, we use calibration activations to transform functional reconstruction into a standard dictionary learning problem. Across LLaMA and Qwen models, CoSpaDi improves accuracy--compression and perplexity--compression trade-offs over SVD-based and structured pruning baselines at 20--40\% compression ratios, while naturally supporting sparse--dense computation and post-training quantization of sparse coefficients.
\end{abstract}
\section{Introduction}

Large language models (LLMs) achieve strong performance across diverse tasks, from dialogue and instruction following~\citep{brown2020language,achiam2023gpt} to general-purpose reasoning~\citep{touvron2023llama,anil2023palm}. Their effectiveness stems in part from transformer architectures that model long-range dependencies with attention~\citep{vaswani2017attention,devlin2019bert}. At the same time, the scale that enables these capabilities makes LLMs expensive to store and run, creating a practical barrier to deployment on memory- and compute-constrained hardware.

A broad literature addresses post-training LLM compression and acceleration, spanning pruning, quantization, knowledge distillation, and matrix factorization~\citep{frankle2018the,dettmers2022gpt3,hinton2015distilling,denton2014exploiting}. Among training-free approaches, matrix factorization is particularly attractive because it yields explicit low-parameter surrogates for large projection matrices. In practice, the predominant choice is truncated SVD and its activation-aware variants, which use a small calibration set to guide which components to keep and how to scale them~\citep{chen2021drone}. Cross-layer extensions further reduce overhead by sharing a common low-dimensional subspace across groups of layers~\citep{wang2025basis}. Despite strong results, these methods ultimately approximate each matrix within a \emph{single} shared low-dimensional subspace; for heterogeneous transformer projections, this constraint can be unnecessarily restrictive and motivates richer factorizations.

In this work, we study an alternative factorization family that relaxes the shared-subspace constraint. Instead of approximating a matrix with one global low-dimensional basis, we model it using a \emph{union of subspaces} via sparse dictionary learning~\citep{aharon2006ksvd,elad2010sparse}: a learned dictionary of atoms is combined with column-sparse coefficients, allowing different columns to be reconstructed from different subsets of atoms. This representation is more flexible than a single shared subspace at the same parameter budget, and it is well aligned with the intuition that different output channels may depend on different latent features.

This sparse parameterization is also timely from a systems perspective. Recent LLM pruning methods show that substantial sparsity can be induced post-training while preserving accuracy~\citep{frantar2023sparsegpt,sun2024wanda}, inference systems increasingly exploit activation and neuron sparsity for efficient serving~\citep{song2024powerinfer}, and specialized sparse kernels have begun to target the moderate-sparsity regime relevant to LLM inference~\citep{macko2025macko}. In parallel, modern accelerator and software stacks expose increasingly mature support for sparse matrix computation, including semi-structured sparsity and sparse tensor abstractions~\citep{mishra2021accelerating,nvidia2020a100,pytorch2024semistructured}. CoSpaDi therefore targets not only a richer approximation class, but also a representation whose coefficient sparsity can be exploited by sparse--dense computation.

Building on this idea, we propose \textbf{CoSpaDi} (\textbf{Co}mpression via \textbf{Spa}rse \textbf{Di}ctionary Learning), a training-free compression framework for transformer projections. CoSpaDi learns dictionaries and sparse codes to approximate pretrained weight matrices, and is \emph{data-aware}: from a small calibration set, we construct an activation-derived Gram orthonormalization that reformulates functional output reconstruction into a standard dictionary learning problem on transformed activation-weighted weights. The resulting factorization yields sparse coefficient representation that can be paired with post-training quantization of the sparse coefficients, and naturally supports \emph{cross-layer} dictionary sharing for groups of related projections.

Overall our \textbf{contributions are three-fold}. \textit{(i)} We introduce sparse dictionary learning as a compression paradigm for LLM weight matrices, replacing the single-subspace constraint of SVD-based factorization with a union-of-subspaces representation.
\textit{(ii)} We integrate sparse dictionary learning with a data-aware objective via activation-derived Gram orthonormalization, yielding a tractable transformed problem that can be optimized with alternating sparse coding and dictionary updates, without gradient-based fine-tuning.
\textit{(iii)} Across multiple LLaMA and Qwen models and a range of compression ratios, CoSpaDi improves the quality--compression trade-off over strong activation-aware SVD baselines in both per-layer and grouped scenarios, and is competitive with recent structured pruning methods.
\begin{figure*}[!t]
    \centering
    \includegraphics[width=0.9\linewidth]{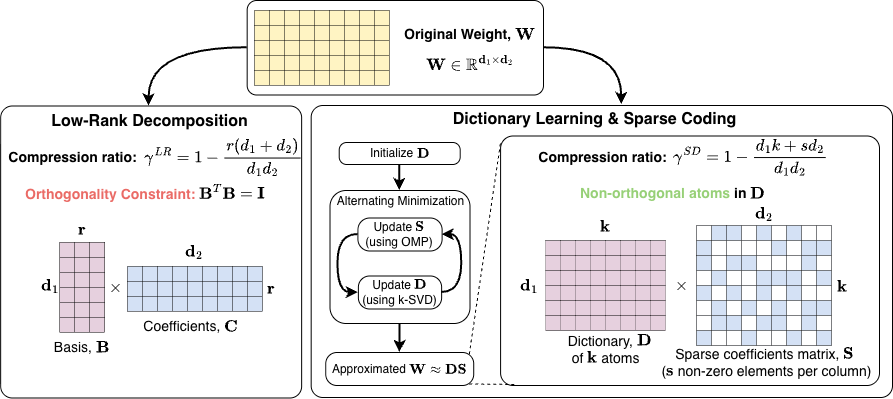}
    \caption{\textit{Left:} low-rank factorization represents $\m W$ in a single $r$-dimensional subspace via two dense factors. \textit{Right:} CoSpaDi represents $\m W$ as $\m D\m S$ where $\m D$ is a dictionary of $k$ atoms and $\m S$ is column-sparse (at most $s$ nonzeros per column), yielding a union-of-subspaces model. Dictionaries may be undercomplete ($k<d_1$), complete ($k=d_1$), or overcomplete ($k>d_1$).}\vspace{-0.5cm}
    \label{fig:proposed_method}
\end{figure*}

\section{Related Work}

\textbf{Low-rank factorization for LLM compression.}
Post-training matrix factorization compresses large neural weights by replacing them with low-parameter surrogates~\citep{denton2014exploiting}. In LLMs, this direction is largely dominated by truncated SVD, which approximates each projection matrix with a rank-$r$ surrogate. Since direct weight-space reconstruction can poorly reflect model behavior, methods such as DRONE~\citep{chen2021drone}, Fisher-weighted reconstruction (FWSVD)~\citep{hsu2022language}, and ASVD~\citep{yuan2023asvd} use calibration data or sensitivity estimates to better preserve layer outputs. More recent SVD-based methods improve this pipeline through truncation-aware whitening and dynamic budget allocation across layers~\citep{wang2025svdllm,wang2025svdllmv2,qinsi2025dobisvd}. Orthogonally, cross-layer approaches such as Basis Sharing reduce storage by reusing low-rank factors across groups of layers~\citep{wang2025basis}. Despite these advances, SVD-based compression retains a common modeling assumption: each matrix is approximated within a \emph{single} shared low-dimensional subspace, possibly shared across layers. CoSpaDi targets the same post-training setting, but replaces this single-subspace model with a union-of-subspaces representation.

\textbf{Sparse dictionary learning and structured factorizations.}
Dictionary learning offers a classical alternative to low-rank subspace models~\citep{engan1999mod,aharon2006ksvd,mairal2009online,gregor2010lista,elad2010sparse}. Instead of expressing all vectors in one shared basis, it learns a dictionary and represents each vector using only a small subset of atoms, yielding a union-of-subspaces model. Beyond sparse coding itself, neural compression has also explored other structured factorization schemes, including block-wise low-rank approximation~\citep{chen2018groupreduce} and tensorized transformer parameterizations~\citep{ma2019tensorized}. Recent transformer applications also use learned dictionaries for other components, such as KV-cache compression via sparse decoding~\citep{kim2025lexico} and cross-layer sharing specialized to attention weights~\citep{zhussip2025shareattention}. In contrast, CoSpaDi develops a training-free, calibration-guided dictionary learning framework for LLM \emph{weight} compression, with both per-layer compression and cross-layer dictionary sharing.

\textbf{Pruning and quantization.}
Pruning and quantization are complementary post-training compression axes. One-shot pruning methods such as SparseGPT~\citep{frantar2023sparsegpt} and Wanda~\citep{sun2024wanda} use calibration data to induce sparsity while preserving layer outputs, while structured pruning removes or replaces higher-level components such as channels, heads, or blocks~\citep{ma2023llmpruner,shopkhoev2025replaceme}. At the systems level, the practical benefit of sparsity depends on metadata overhead, memory access patterns, sparsity structure, and kernel or hardware support~\citep{han2015deepcompression,wang2020sparsert,mishra2021accelerating,nvidia2020a100}. CoSpaDi is related to this line in that it introduces sparsity in the coefficient matrix of a learned factorization, so its runtime benefits are subject to similar implementation considerations. Quantization reduces numerical precision and addresses outliers or activation imbalance through Hessian-aware reconstruction, activation-aware scaling, sparse outlier handling, or rotations~\citep{dettmers2022gpt3,frantar2023gptq,frantar2023optq,lin2024awq,dettmers2024spqr,xiao2023smoothquant,ashkboos2024quarot}. These techniques are orthogonal to CoSpaDi, and can be applied post hoc to the sparse coefficients matrix.
\vspace{-0.1cm}\section{Method}\vspace{-0.2cm}
\label{sec:method}

CoSpaDi compresses pretrained transformer projections by replacing each dense weight matrix with a dense dictionary and sparse column-wise coefficients. 
We first formulate the activation-space objective used for calibration-guided compression, then reinterpret SVD as a basis--coefficient model, and finally introduce sparse dictionary learning as a union-of-subspaces alternative.

\subsection{Activation-space reconstruction}
\label{sec:problem_formulation}

Consider a pretrained projection $\m W\in\R^{d_1\times d_2}$ with calibration activations $\m X\in\R^{N\times d_1}$ and outputs $\m X\m W$.
Post-training compression replaces $\m W$ by a structured approximation $\tilde{\m W}\in\mathcal{C}$, where $\mathcal{C}$ denotes the chosen compressed family.
Instead of minimizing weight-space error $\|\m W-\tilde{\m W}\|_F^2$, we preserve the induced layer outputs:
\begin{equation}
    \tilde{\m W}^{\star}
    =
    \argmin_{\tilde{\m W}\in\mathcal{C}}
    \|\m X\m W-\m X\tilde{\m W}\|_F^2 .
    \label{eq:activation_objective}
\end{equation}
This objective is used implicitly or explicitly in many calibration-guided post-training methods and motivates the activation-aware design of CoSpaDi.

To remove explicit activations from the optimization, let $\m \Delta=\m W-\tilde{\m W}$ and $\m G=\m X^\transp\m X$ be the uncentered activation Gram matrix.
We compute a numerically stabilized Gram factor $\m L$ such that $\m L^\top\m L$ approximates $\m G$ (See~\ref{apx:data_aware} for details).
Then
\begin{equation}
\begin{aligned}
    \|\m X\m \Delta\|_F^2
    &=
    \mathrm{tr}\!\left(
    \m \Delta^\transp \m X^\transp\m X \m \Delta
    \right) \\
    &=
    \mathrm{tr}\!\left(
    \m \Delta^\transp \m G \m \Delta
    \right)
    \approx
    \|\m L\m \Delta\|_F^2 .
\end{aligned}
\label{eq:activation_to_weight}
\end{equation}

Thus, Eq.~\eqref{eq:activation_objective} reduces to Frobenius reconstruction of the transformed weights $\m W_L=\m L\m W$.

\subsection{From SVD to sparse dictionary learning}
\label{sec:svd_to_dictionary}

Low-rank compression restricts the approximation to matrices of rank at most $r$:
\begin{equation}
    \tilde{\m W}^{\mathrm{LR}}_L
    =
    \argmin_{\mathrm{rank}(\tilde{\m W}_L)\le r}
    \|\m W_L-\tilde{\m W}_L\|_F^2 .
    \label{eq:low_rank_approx}
\end{equation}
By the Eckart--Young--Mirsky theorem~\citep{eckart1936approximation}, the solution is the truncated SVD
$\tilde{\m W}^{\mathrm{LR}}_L=\m U_r\bm\Sigma_r\m V_r^\transp$, where $(\m U_r,\bm\Sigma_r,\m V_r)$ are the top-$r$ singular components of $\m W_L$.

This solution also admits the standard PCA projection interpretation~\citep{bishop2006pattern}. 
Viewing the columns of $\m W_L$ as vectors in $\R^{d_1}$, SVD searches for an $r$-dimensional subspace that minimizes their reconstruction error. 
Equivalently, it solves the basis--coefficient problem
\begin{equation}
    \min_{\m B,\m C}
    \|\m W_L-\m B\m C\|_F^2
    \quad
    \mathrm{s.t.}
    \quad
    \m B^\transp\m B=\m I ,
    \label{eq:pca_loss}
\end{equation}
with optimum $\m B^\star=\m U_r$ and $\m C^\star=\bm\Sigma_r\m V_r^\transp$; see Appendix~\ref{apx:PCA_proof}. 
Thus, truncated SVD is a basis--coefficient model in which every column is reconstructed in the same shared subspace $\mathrm{span}(\m B)$, while only its dense coordinates vary.

Sparse dictionary learning keeps the basis--coefficient form but changes both modeling constraints. 
The orthonormal basis $\m B$ is replaced by a generally non-orthogonal dictionary $\m D_L\in\R^{d_1\times k}$, and the dense coefficient matrix $\m C$ is replaced by a column-sparse coefficient matrix $\m S\in\R^{k\times d_2}$:
\begin{equation}
    \min_{\m D_L,\m S}
    \|\m W_L-\m D_L\m S\|_F^2
    \quad
    \mathrm{s.t.}
    \quad
    \|\m s_j\|_0\le s,\quad \forall j .
    \label{eq:DicLearn_data_aware_final}
\end{equation}
For column $j$, the support $\mathcal{T}_j=\mathrm{supp}(\m s_j)$ selects the active atoms (columns), so the weight column is reconstructed in $\mathrm{span}(\m D_{L,\mathcal{T}_j})$. 
Since different columns may choose different supports, CoSpaDi replaces the single shared SVD subspace with a union of at most $s$-dimensional subspaces. 
After solving Eq.~\eqref{eq:DicLearn_data_aware_final}, we map back to the original parameterization:
\begin{equation}
    \tilde{\m W}
    =
    \m D_a\m S,
    \qquad
    \m D_a=\m L^{-1}\m D_L .
    \label{eq:map_back}
\end{equation}

\subsection{Optimization and grouped sharing}
\label{sec:optimization_grouping}

The objective in Eq.~\eqref{eq:DicLearn_data_aware_final} is non-convex due to the bilinear factorization and sparsity constraints. 
We optimize it by alternating between sparse coding and dictionary update. 
With $\m D_L$ fixed, each column is encoded independently:
\begin{equation}
    \m s_j
    \leftarrow
    \argmin_{\|\m s\|_0\le s}
    \|\m w_{L,j}-\m D_L\m s\|_2^2 ,
    \label{eq:sparse_coding}
\end{equation}
which we approximate with batched orthogonal matching pursuit (OMP)~\citep{tropp2007omp,rubinstein2008efficient}. 
With $\m S$ fixed, the dictionary is updated by
\begin{equation}
    \m D_L
    \leftarrow
    \argmin_{\m D}
    \|\m W_L-\m D\m S\|_F^2 .
    \label{eq:dictionary_update}
\end{equation}
This step can be performed by MOD, which updates all atoms jointly~\citep{engan1999mod}, or by K-SVD, which updates atoms sequentially via rank-one residual approximations~\citep{aharon2006ksvd}.

Because Eq.~\eqref{eq:DicLearn_data_aware_final} is non-convex, we do not claim global optimality. Under exact sparse-coding and dictionary block minimization, however, the reconstruction objective is monotonically non-increasing and, being bounded below, converges to a finite value. Our practical OMP and K-SVD/power-iteration updates approximate these block solutions; we discuss the corresponding guarantees and their limitations in Appendix~\ref{apx:convergence}.
The full optimization procedure is summarized in \ref{apx:algo}.

CoSpaDi also supports sharing one dictionary across related projections (see \ref{apx:grouped}). 
For a group $\mathcal{G}=\{\ell_1,\ldots,\ell_m\}$, we concatenate weights as
$\m W_{\mathcal{G}}=[\m W_{\ell_1},\ldots,\m W_{\ell_m}]$ and vertically stack calibration inputs to form $\m X_{\mathcal{G}}$. 
We compute the Gram factor from $\m X_{\mathcal{G}}$, solve Eq.~\eqref{eq:DicLearn_data_aware_final} on $\m W_{\mathcal{G},L}=\m L\m W_{\mathcal{G}}$, and then slice the learned coefficient matrix into layer-specific blocks. 
This amortizes the dictionary cost across layers while preserving layer-specific sparse coefficients, following the motivation of cross-layer basis sharing~\citep{wang2025basis}.

\subsection{Storage and inference cost}
\label{sec:storage_complexity}

We report the compression ratio as storage reduction relative to the dense matrix. 
For low-rank factorization, storing two dense factors costs $r(d_1+d_2)$ values:
\begin{equation}
    \gamma^{\mathrm{LR}}
    =
    1-\frac{r(d_1+d_2)}{d_1d_2}.
    \label{eq:cr_lr}
\end{equation}
For CoSpaDi, the dictionary costs $d_1k$ values and the sparse coefficients contain $sd_2$ nonzero values. 
With $\rho=k/s$, a packed binary mask for the coefficient support costs $kd_2=\rho sd_2$ bits, or $\rho sd_2/16$ bfloat16 words:
\begin{equation}
    \gamma^{\mathrm{SD}}_{\mathrm{mask}}
    =
    1-
    \frac{
        d_1k+sd_2+\frac{\rho sd_2}{16}
    }{
        d_1d_2
    } .
    \label{eq:cr_sd_mask}
\end{equation}
In the main experiments, we absorb this mask cost by storing each nonzero coefficient with $\rho$ fewer bits: saving $\rho$ bits for each of the $sd_2$ coefficients exactly offsets the $\rho sd_2$ mask bits. 
This gives the effective accounting
\begin{equation}
    \gamma^{\mathrm{SD}}
    =
    1-
    \frac{
        d_1k+sd_2
    }{
        d_1d_2
    } .
    \label{eq:cr_sd_effective}
\end{equation}

Thus, once the target compression ratio $\gamma^{\mathrm{SD}}$ and allocation ratio $\rho=k/s$ are fixed, both the dictionary size $k$ and sparsity level $s$ are uniquely determined by the storage budget.

For $N$ input tokens, dense inference costs $Nd_1d_2$ multiplications, while low-rank inference costs $Nr(d_1+d_2)$. 
CoSpaDi evaluates $\m X\tilde{\m W}=(\m X\m D_a)\m S$; with reuse of inner products over the active dictionary atoms, its sparse cost is $Nd_1K_{\mathrm{active}}+Nsd_2$, where $K_{\mathrm{active}}$ is the number of dictionary atoms used by at least one column. 
Actual latency depends on the sparsity pattern, mask handling, memory traffic, and sparse--dense kernel efficiency. We provide the full storage accounting, theoretical complexity derivations, and empirical inference measurements in~\ref{apx:cr}, ~\ref{apx:inference_complexity} and ~\ref{apx:deployment}, respectively.
\begin{figure*}[!t]
    \centering
    \begin{minipage}{0.53\textwidth}
        \centering
        \includegraphics[width=\textwidth]{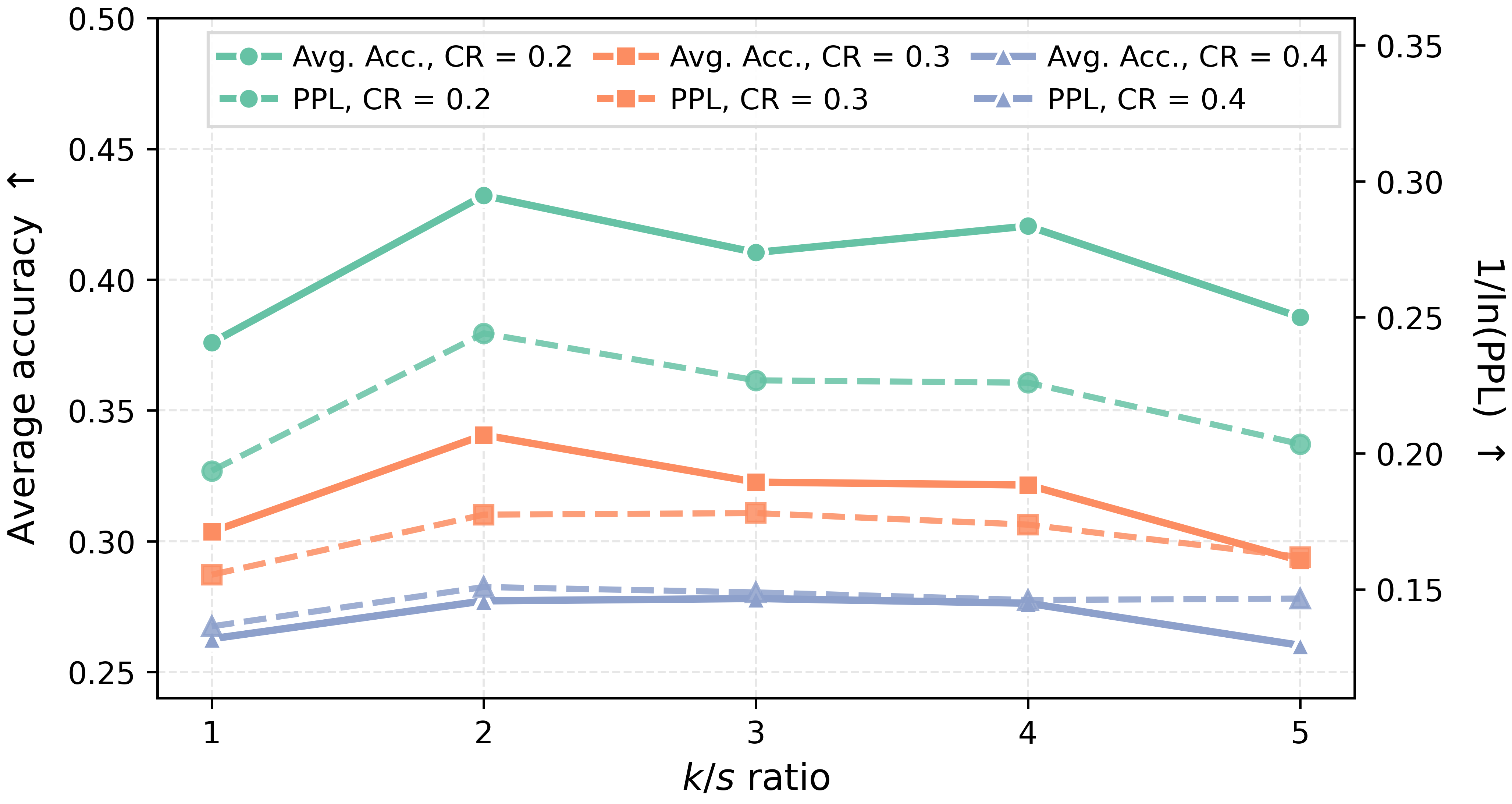}
        \captionof{figure}{Dual-axis plot showing average accuracy (\solidline{black} solid lines, left axis) and WikiText perplexity (\dashedline{black} dashed lines, right axis, inverted logarithmic scale) as functions of $\rho$ for LLaMA-3.2-1B under three CRs: \textcolor{teal}{0.2}, \textcolor{coral}{0.3} and \textcolor{lavender}{0.4}.}
        \label{fig:ablation_ks}

        \includegraphics[width=0.8\textwidth]{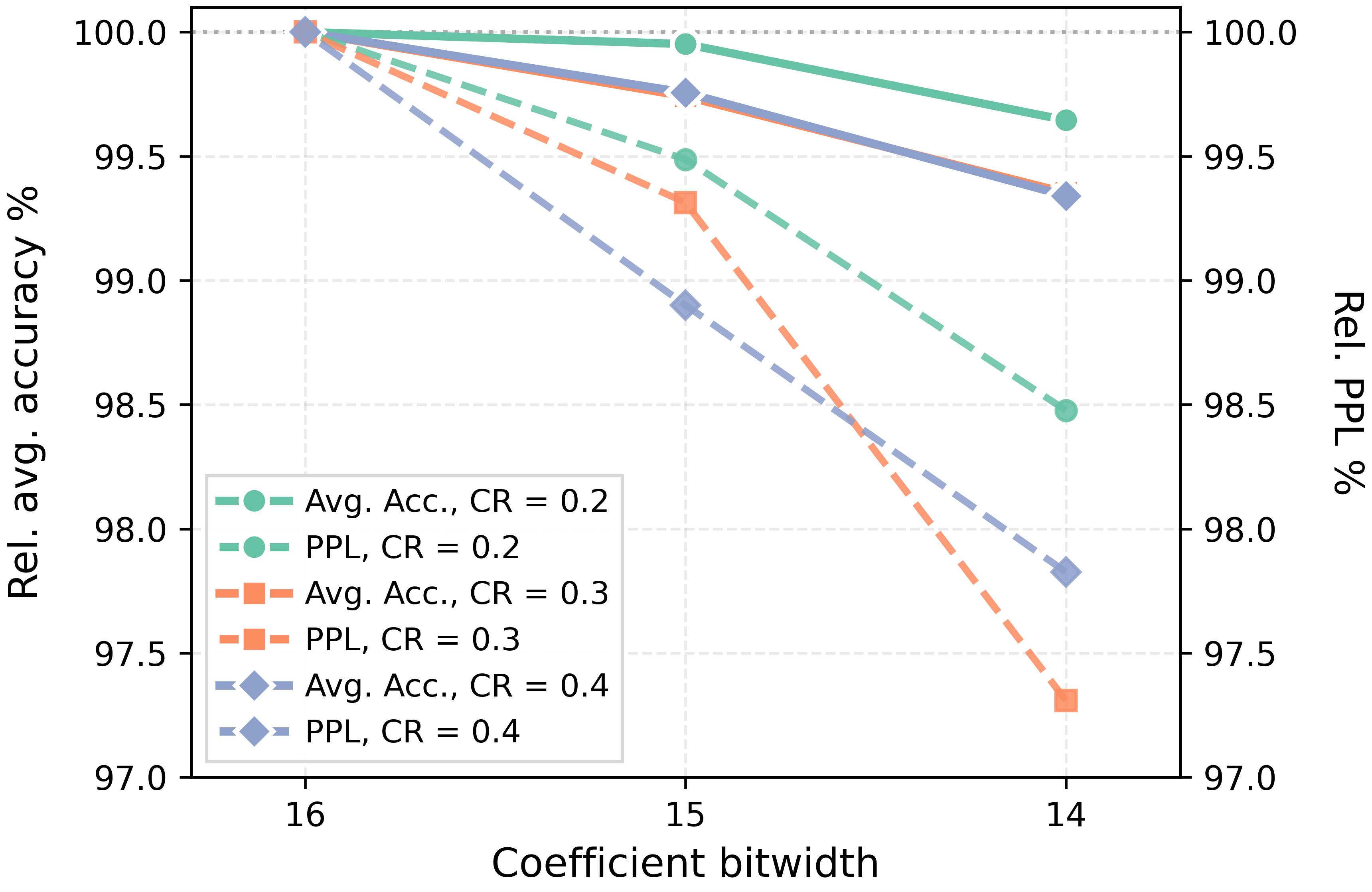}
        \captionof{figure}{Dual-axis plot showing relative average accuracy (\solidline{black} solid lines, left axis) and relative WikiText perplexity (\dashedline{black} dashed lines, right axis) as functions of $\m S$ bitwidth for LLaMA-3-8B under three CRs: \textcolor{teal}{0.2}, \textcolor{coral}{0.3} and \textcolor{lavender}{0.4}.} \vspace{-0.5cm}
        \label{fig:ablation_quant}
    \end{minipage}%
    \begin{minipage}{0.02\textwidth}
     \hfill
    \end{minipage}%
    \begin{minipage}{0.45\textwidth}

        \resizebox{\textwidth}{!}{%
        \renewcommand{\arraystretch}{1.05}
        \begin{tabular}{cccccc}
        \hline
         & & & & \multicolumn{2}{c}{\textbf{Perplexity$\downarrow$}} \\ \cline{5-6}
        \multirow{-2}{*}{\textbf{Method}} & \multirow{-2}{*}{\textbf{Data}} & \multirow{-2}{*}{\textbf{CR}} & \multirow{-2}{*}{\textbf{Avg. Acc.$\uparrow$}} & \textbf{WikiText} & \textbf{LAMBADA}    \\ \hline
        \textbf{LLaMA-3.2-1B} & -- & --              & 57.6 & 11.6E+00 & 5.7E+00 \\ \hline \hline
        
        SVD               & \xmark   &              & 23.9 & 2.9E+06 & 4.6E+06          \\
        $\text{CoSpaDi} ^ \dag$    & \xmark   &     & \textbf{24.5} & \textbf{3.3E+05} & \textbf{2.2E+06}          \\ \cdashline{4-6}
        SVD-LLM  & \cmark   &                       & 37.6 & 1.7E+02 & 1.7E+02          \\
        CoSpaDi  & \cmark   & \multirow{-4}{*}{0.2} & \textbf{42.7} & \textbf{6.4E+01} & \textbf{3.5E+01} \\ \hline
        
        SVD      & \xmark   &                       & 24.3 & 1.1E+06 & \textbf{3.9E+06}          \\
        $\text{CoSpaDi} ^ \dag$    & \xmark   &     & \textbf{24.5} & \textbf{2.1E+05} & 4.3E+06          \\ \cdashline{4-6}
        SVD-LLM  & \cmark   &                       & 30.2          & 5.9E+02 & 2.5E+03          \\
        CoSpaDi  & \cmark   & \multirow{-4}{*}{0.3} & \textbf{33.7} & \textbf{2.9E+02} & \textbf{6.6E+02} \\ \hline
        
        SVD      & \xmark   &                       & 24.1          & \textbf{1.2E+06}          & \textbf{4.2E+06}          \\
        $\text{CoSpaDi} ^ \dag$    & \xmark   &     & \textbf{24.7} & 3.1E+06          & 3.7E+07          \\ \cdashline{4-6}
        SVD-LLM  & \cmark   &                       & 26.2 & 1.6E+03          & 3.3E+04          \\
        CoSpaDi  & \cmark   & \multirow{-4}{*}{0.4} & \textbf{27.5} & \textbf{8.0E+02} & \textbf{9.2E+03} \\ \hline
        \end{tabular}
        }
        \captionof{table}{SDL-based methods comparison vs low-rank counterparts in data-free and data-aware scenarios on LLaMA-3.2-1B at different CR. We denote $\text{CoSpaDi} ^ \dag$ as the proposed method without using calibration data. Best results are provided in \textbf{bold}.}
        \label{tab:ablation_data}

        \vspace{0.7cm}
        
        \resizebox{\textwidth}{!}{%
        \renewcommand{\arraystretch}{1.35}
        \begin{tabular}{ccccc}
        \hline
         & \textbf{Wall-clock} &  & \multicolumn{2}{c}{\textbf{Perplexity$\downarrow$}} \\ \cline{4-5}
        \multirow{-2}{*}{\textbf{Method}} & \textbf{time} & \multirow{-2}{*}{\textbf{Avg. Acc.$\uparrow$}} & \textbf{WikiText} & \textbf{LAMBADA}    \\ \hline
        \textbf{LLaMA-3.2-1B} & -- & 57.6 & 11.6E+00 & 5.7E+00 \\ \hline \hline
        
        MOD             & \textbf{222.2} & 38.9 & 1.3E+02 & 1.1E+02 \\ 
        K-SVD (PCA)     & 902.1 & 41.5 & 9.2E+01 & 5.4E+01  \\ 
        K-SVD (power)   & 646.8 & \textbf{42.7} & \textbf{6.4E+01} & \textbf{3.5E+01} \\  
        \hline
        \end{tabular}
        }
        \captionof{table}{Solver ablation for CoSpaDi dictionary updates on LLaMA-3.2-1B at 0.2 CR, $\rho = 2$ with 60 alternating minimization iterations. We report wall-clock compression time in minutes using a single A100. Best results are provided in \textbf{bold}.} \vspace{-0.5cm}
        \label{tab:abl_solvers}
        
    \end{minipage}
\end{figure*}
\begin{table*}[h]
\resizebox{\textwidth}{!}{%
\renewcommand{\arraystretch}{1.05}
\begin{tabular}{ccccccccccc:cc}
\hline
                            &                         & \multicolumn{9}{c}{\textbf{Accuracy$\uparrow$}}                                                                           & \multicolumn{2}{c}{\textbf{Perplexity$\downarrow$}}       \\ \cline{3-13}
\multirow{-2}{*}{\textbf{Method}}    & \multirow{-2}{*}{\textbf{CR}}    & \textbf{PIQA} & \textbf{HellaSwag} & \textbf{LAMBADA} & \textbf{ARC-e} & \textbf{ARC-c} & \textbf{SciQ} & \textbf{RACE}& \textbf{MMLU} & \textbf{Avg.} & \textbf{WikiText} & \textbf{LAMBADA} \\ \hline

\textbf{LLaMA-3-8B}        & --                      & 80.7 & 79.1 & 75.6 & 77.7 & 53.5 & 93.9 & 40.3 & 62.2 & 70.4 & 7.3E+00 & 3.1E+00 \\ \hline \hline

SVD-LLM                     &                         & 71.1 & 58.4 & 59.3 & 55.5 & 34.0 & 86.4 & 35.5 & 32.6 & 54.1 & 4.1E+01 & 1.1E+01 \\
CoSpaDi                     & \multirow{-2}{*}{0.2}   & 75.2 & 66.5 & 73.8 & 66.5 & 41.6 & 89.5 & 38.2 & 42.8 & \textbf{61.8} & \textbf{2.0E+01} & \textbf{4.3E+0}0 \\ \hline

SVD-LLM                     &                         & 65.8 & 46.4 & 38.1 & 41.9 & 27.7 & 70.0 & 31.8 & 27.2 & 43.6 & 1.5E+02 & 6.1E+01 \\
CoSpaDi                     & \multirow{-2}{*}{0.3}   & 70.5 & 56.2 & 61.3 & 54.2 & 33.5 & 85.7 & 36.2 & 32.2 & \textbf{53.7} & \textbf{4.5E+01} & \textbf{9.2E+00} \\ \hline

SVD-LLM                     &                         & 60.3 & 34.5 & 11.4 & 32.4 & 24.5 & 44.2 & 25.7 & 23.1 & 32.0 & 5.5E+02 & 1.3E+03 \\
CoSpaDi                     & \multirow{-2}{*}{0.4}   & 63.7 & 41.4 & 30.3 & 39.1 & 26.6 & 68.5 & 30.5 & 25.4 & \textbf{40.7} & \textbf{1.8E+02} & \textbf{1.2E+02} \\ \hline

\textbf{Qwen3-8B}          & --                      & 77.7 & 74.9 & 64.1 & 80.7 & 56.7 & 95.7 & 40.9 & 73.0 & 70.5 & 1.2E+01 & 4.6E+00 \\ \hline \hline

SVD-LLM                    &                         & 73.8 & 63.9 & 62.2 & 68.7 & 45.7 & 90.1 & 40.5 & 54.7 & 62.5 & 2.1E+01 & 6.4E+00 \\
CoSpaDi                    & \multirow{-2}{*}{0.2}   & 76.5 & 68.0 & 65.6 & 72.2 & 48.9 & 93.2 & 40.7 & 60.8 & \textbf{65.7} & \textbf{1.8E+01} & \textbf{4.9E+00} \\ \hline

SVD-LLM                    &                         & 70.4 & 55.2 & 53.8 & 59.3 & 37.1 & 87.2 & 38.4 & 44.8 & 55.8 & 2.7E+01 & 1.1E+01 \\
CoSpaDi                    & \multirow{-2}{*}{0.3}   & 72.4 & 60.5 & 62.6 & 63.9 & 41.2 & 88.4 & 39.5 & 51.3 & \textbf{59.97} & \textbf{2.3E+01} & \textbf{6.3E+00} \\ \hline

SVD-LLM                    &                         & 66.3 & 44.6 & 37.9 & 45.0 & 28.1 & 77.3 & 35.3 & 29.1 & 45.4 & 4.3E+01 & 3.6E+01 \\
CoSpaDi                    & \multirow{-2}{*}{0.4}   & 68.9 & 49.0 & 49.9 & 49.4 & 29.9 & 82.0 & 36.8 & 36.6 & \textbf{50.3} & \textbf{3.6E+01} & \textbf{1.5E+01} \\ \hline

\end{tabular}
}
\caption{Performance comparison of CoSpaDi and SVD-LLM on LLaMA-3-8B and Qwen3-8B across compression ratios and evaluation benchmarks. Best results are provided in \textbf{bold}.}
\label{tab:main_results}
\end{table*}
\begin{table*}[h]
\resizebox{\textwidth}{!}{%
\renewcommand{\arraystretch}{1.05}
\begin{tabular}{ccccccccccc:cc}
\hline
                              &                                           & \multicolumn{9}{c}{\textbf{Accuracy$\uparrow$}}                                                                           & \multicolumn{2}{c}{\textbf{Perplexity$\downarrow$}}       \\ \cline{3-13}
\multirow{-2}{*}{Method}      & \multirow{-2}{*}{CR}                      & \textbf{PIQA} & \textbf{HellaSwag} & \textbf{LAMBADA} & \textbf{ARC-e} & \textbf{ARC-c} & \textbf{SciQ} & \textbf{RACE}& \textbf{MMLU} & \textbf{Avg.} & \textbf{WikiText} & \textbf{LAMBADA} \\ \hline
 
\multicolumn{2}{c}{\textbf{LLaMA-3-8B}}                & 80.7 & 79.1 & 75.6 & 77.7 & 53.5 & 93.9 & 40.3 & 62.2 & 70.4    & 7.3E+00       & 3.1E+00            \\ \hline \hline

ReplaceMe                   & 0.22                    & 73.1 & 65.7 & 42.1 & 65.9 & 43.7 & 86.4 & 35.4 & 51.7 & 58.0    & 3.4E+01       & 2.0E+01            \\
LLM-Pruner                  &                         & 75.5 & 67.5 & 51.0 & 62.1 & 36.6 & 87.8 & 35.1 & 25.0 & 55.1    & \textbf{1.6E+01}       & 1.1E+01            \\
CoSpaDi                     & \multirow{-2}{*}{0.2}   & 75.2 & 66.5 & 73.8 & 66.5 & 41.6 & 89.5 & 38.2 & 42.8 & \textbf{61.8}    & 2.0E+01       & \textbf{4.3E+00} \\ \hline

ReplaceMe                   & 0.31                    & 66.6 & 53.8 & 24.0 & 50.7 & 37.9 & 77.3 & 34.0 & 30.6 & 46.9    & 6.7E+01       & 1.3E+02            \\
LLM-Pruner                  &                         & 67.3 & 45.1 & 20.9 & 45.4 & 28.8 & 63.4 & 30.1 & 22.9 & 40.5    & \textbf{3.8E+01}       & 2.2E+02            \\
CoSpaDi                     & \multirow{-2}{*}{0.3}   & 70.5 & 56.2 & 61.3 & 54.2 & 33.5 & 85.7 & 36.2 & 32.2 & \textbf{53.7}    & 4.5E+01       & \textbf{9.2E+00} \\ \hline

ReplaceMe                   & 0.41                    & 61.7 & 44.3 & 9.8  & 37.4 & 27.5 & 60.4 & 31.6 & 26.4 & 37.4    & 2.3E+02       & 1.8E+03            \\
LLM-Pruner                  &                         & 50.3 & 25.8 & 1.5  & 26.4 & 25.8 & 28.1 & 21.8 & 23.2 & 25.4    & $\infty$       & 5.7E+05            \\
CoSpaDi                     & \multirow{-2}{*}{0.4}   & 63.7 & 41.4 & 30.3 & 39.1 & 26.6 & 68.5 & 30.5 & 25.4 & \textbf{40.7}    & \textbf{1.8E+02}       & \textbf{1.2E+02} \\ \hline

\end{tabular}
}
\caption{Comparison of the proposed CoSpaDi method with state-of-the-art structured pruning methods ReplaceMe \cite{shopkhoev2025replaceme} and LLM-Pruner \cite{ma2023llmpruner} on LLaMA-3-8B under different compression ratios. We report accuracy on different benchmarks as well as its average and perplexity. Best results are provided in \textbf{bold}.} \vspace{-0.5cm}
\label{tab:main_llama3_8b_train_free}
\end{table*}

\vspace{-0.1cm}\section{Experiments}
\label{sec:experiments}
We evaluate CoSpaDi in two regimes: \textbf{per-layer} compression, where each projection matrix is compressed independently, and \textbf{grouped (cross-layer)} compression, where a dictionary is shared across a set of layers of the same type. We first present ablations that isolate key design choices (dictionary capacity allocation, data-awareness, packing/quantization, and solver variants), and then report main results for both regimes.

\vspace{-0.2cm}\subsection{Experimental Setup}
\label{sec:exp_setup}
\paragraph{Models and layers.}
We evaluate per-layer compression on \texttt{LLaMA-3.2-1B}, \texttt{Qwen3-0.6B}, \texttt{LLaMA-3-8B}, \texttt{Qwen3-8B} and \texttt{Qwen3-14B}.
For \emph{SVD-LLM} we use the authors' original implementation\footnote{\url{https://github.com/AIoT-MLSys-Lab/SVD-LLM}} and perform
only the post-training compression stage, without the optional
fine-tuning step.
For grouped compression, we follow the \emph{Basis Sharing}\footnote{\url{https://github.com/TUDa-HWAI/Basis_Sharing}} protocol and report results on \texttt{LLaMA-2-7B}.
Unless stated otherwise, we compress all dense linear projections in self-attention (Q/K/V/O) and gated MLP (up/down/gate), while leaving embeddings and the LM head intact.

\paragraph{Calibration data and metrics.}
For calibration we randomly sample 256 sequences of length 1024 from RefinedWeb \citep{penedo2023the}. We study sensitivity to the calibration domain and the number of calibration samples in Appendix~\ref{apx:calibration_sensitivity}.
We report standard zero-shot accuracy (normalized when available) on a suite of benchmarks (PIQA \citep{bisk2020piqa}, HellaSwag \citep{zellers2019hellaswag}, OpenAI LAMBADA \citep{paperno2016lambada}, ARC-easy/ARC-challenge \citep{clark2018think}, SciQ \citep{welbl2017crowdsourcing}, RACE \citep{lai2017race}, MMLU \citep{hendryckstest2021}) and perplexity on WikiText \citep{merity2017pointer} and OpenAI LAMBADA. We used lm-evaluation-harness 0.4.8 \citep{eval-harness} to ensure reproducibility. We additionally use task-level standard errors from
\texttt{lm-evaluation-harness} to analyze the uncertainty of the
CoSpaDi--SVD-LLM average-accuracy gap in Appendix~\ref{apx:evaluation_uncertainty}.

\textbf{Baselines.}
In the per-layer setting, we compare primarily to SVD-LLM~\citep{wang2025svdllm}, a strong activation-aware SVD baseline, and, where relevant, to plain truncated SVD as its data-free counterpart. This comparison keeps the calibration protocol and 
structural storage budget matched, allowing us to isolate the effect of replacing single-subspace with union-of-subspaces factorization.

More recent extensions such as SVD-LLM V2~\citep{wang2025svdllmv2} and
Dobi-SVD~\citep{qinsi2025dobisvd} additionally introduce heterogeneous rank allocation,
loss-aware truncation, differentiable rank selection, and low-bit remapping.
We discuss these differences and provide additional empirical comparisons in
Appendix~\ref{apx:orthogonal_svd_extensions}. These methods address complementary
aspects of the compression pipeline that can in principle also be combined with
CoSpaDi.

\textbf{Reproducibility details.}
For reproducibility, we provide a consolidated summary of the implementation details used in the main experiments in~\ref{apx:impl_details}, including the solver configuration, dictionary initialization, sparse-coding setup, calibration protocol
and a table of the main hyperparameters.

\subsection{Ablation Studies}
\label{sec:ablations}
\textbf{Capacity Allocation: the \texorpdfstring{$k/s$}{k/s} Ratio.}
\label{sec:ablation_rho}
At a fixed compression ratio (CR), CoSpaDi admits a family of factorizations parameterized by $\rho \coloneqq k/s$.
Smaller $\rho$ increases per-column expressiveness (larger $s$) but reduces the dictionary size $k$; larger $\rho$ increases $k$ but restricts each column to fewer active atoms.
We perform a coarse sweep over $\rho \in [1;5]$.
Figure~\ref{fig:ablation_ks} (LLaMA-3.2-1B) shows that $\rho=2$ consistently provides the best trade-off across all considered CRs in terms of both average accuracy and perplexity, so we fix $\rho=2$ in all subsequent experiments.

\textbf{Data-Free vs.\ Data-Aware Compression}
\label{sec:ablation_dataaware}
A core motivation for dictionary learning is that its union-of-subspaces form can be more expressive than a single low-dimensional subspace \emph{regardless of whether calibration is used}.
We therefore compare:
(i) \textbf{data-free} truncated SVD vs.\ CoSpaDi$^\dagger$ (dictionary learning in weight space, without whitening), and
(ii) \textbf{data-aware} SVD-LLM vs.\ CoSpaDi (activation-aware via whitening).
Table~\ref{tab:ablation_data} summarizes results on LLaMA-3.2-1B across CR 0.2--0.4 reporting average accuracy and perplexities.
In both regimes, the dictionary-based factorization is consistently stronger than the corresponding low-rank baseline, and activation-aware CoSpaDi yields the best overall trade-off.

\textbf{Dictionary Learning Solver Variants.}
\label{sec:ablation_solvers}
The main computational bottleneck in CoSpaDi is the dictionary update: K-SVD updates atoms sequentially, which can be expensive for large projection matrices. We therefore compare several dictionary-update variants, including exact rank-1 updates via truncated PCA, approximate rank-1 updates via power iterations, and MOD, which updates the whole dictionary in a single least-squares step.

Table~\ref{tab:abl_solvers} reports quality and wall-clock compression time on LLaMA-3.2-1B at 0.2 CR, using the same hardware, batching, stopping criteria, and fixed factorization shape $(k,s)$. This isolates the solver-level speed--quality trade-off. We use power-iteration K-SVD in the main experiments, as it provides the best overall balance between accuracy, perplexity, and compression time.
Additional ablations over the number of alternating steps and power iterations are provided in~\ref{apx:iter_ablation}, and~\ref{apx:runtime_breakdown} further breaks down the wall-clock cost into sparse coding, dictionary updates, and miscellaneous overhead, confirming that the sequential dictionary-update stage dominates runtime.
In practice, compression is naturally parallelizable across independent projection matrices. For our model-scale experiments, we use an $8\times$A100 node with two independent compression processes per GPU. This provides up to 16-way task parallelism and, under ideal workload balancing, can reduce wall-clock time by up to $16\times$.

\textbf{Coefficient Quantization}
\label{sec:ablation_quant}

Sparse factorization requires storing both nonzero coefficient values and their locations. 
We therefore study post-training coefficient quantization by truncating bf16 mantissa bits. 
As shown in Figure~\ref{fig:ablation_quant}, with our default $\rho=k/s=2$, truncating 2 mantissa bits yields almost negligible degradation across tested CRs. 
This truncation offsets the packed mask overhead, matching the effective storage accounting in Eq.~\ref{eq:cr_sd_effective}. 
Unless stated otherwise, we evaluate CoSpaDi with bf16 dictionaries and 14-bit sparse coefficients. 
Appendix~\ref{apx:storage_breakdown} provides a component-wise memory breakdown.

\begin{table*}[h]
\resizebox{\textwidth}{!}{%
\renewcommand{\arraystretch}{1.05}
\begin{tabular}{lcccccccccc:cc}
\hline
\multirow{2}{*}{\textbf{Method}} & \multirow{2}{*}{\textbf{CR}} & \multicolumn{9}{c}{\textbf{Accuracy$\uparrow$ (\% (unnorm))}} & \multicolumn{2}{c}{\textbf{Perplexity$\downarrow$}} \\ \cline{3-13}
                                  &                             & \textbf{PIQA} & \textbf{HellaSwag} & \textbf{LAMBADA} & \textbf{ARC-e} & \textbf{ARC-c} & \textbf{SciQ} & \textbf{RACE} & \textbf{MMLU} & \textbf{Avg.} & \textbf{WikiText} & \textbf{LAMBADA} \\ \hline

\textbf{LLaMA-2-7B}                & --                          & 78.9 (78.0) & 76.1 (57.3) & 73.8 & 74.2 (76.0) & 45.8 (42.7) & 91.4 (93.9) & 39.7 & 40.8 & 65.1 (62.8) & 8.70 & 3.38 \\ \hline \hline

SVD-LLM                           & \multirow{4}{*}{0.2}        & 73.1 (72.1) & 60.3 (44.4) & 63.9 & 55.0 (62.0) & 32.0 (29.1) & 79.5 (89.2) & 36.6 & 25.5 & 53.2 (52.9) & 20.18 & 6.33 \\

CoSpaDi (per-layer)               &                             & 74.5 (73.6) & 64.5 (47.7) & 70.0 & 61.5 (67.2) & 35.2 (34.0) & 85.4 (91.3) & 39.2 & 28.0 & 57.3 (56.4) & 15.97 & 4.58 \\ \cdashline{3-13}
Basis Sharing                     &                             & 71.1 (70.2) & 60.0 (43.4) & 62.8 & 60.5 (65.5) & 37.8 (33.5) & 85.0 (90.8) & 34.7 & 25.0 & 54.6 (53.2) & 15.17 & 7.03 \\

CoSpaDi (grouped)                 &                             & 75.3 (74.5) & 66.3 (48.8) & 71.1 & 68.1 (71.3) & 39.3 (37.3) & 88.5 (92.2) & 38.5 & 27.0 & \textbf{59.3 (57.6)} & \textbf{11.73} & \textbf{4.42} \\ \hline

SVD-LLM                           & \multirow{4}{*}{0.3}        & 68.2 (68.0) & 52.2 (39.4) & 51.6 & 47.3 (53.6) & 28.1 (25.2) & 75.9 (84.9) & 34.6 & 23.5 & 47.7 (47.6) & 33.99 & 13.54 \\

CoSpaDi (per-layer)               &                             & 71.4 (70.1) & 57.1 (42.4) & 62.7 & 53.2 (60.2) & 31.1 (27.5) & 81.7 (88.3) & 36.3 & 26.7 & 52.5 (51.8) & 23.90 & 6.75 \\ \cdashline{3-13}
Basis Sharing                     &                             & 66.5 (65.4) & 50.3 (37.6) & 53.6 & 54.2 (58.7) & 29.3 (27.2) & 81.4 (86.9) & 32.4 & 23.3 & 48.9 (48.1) & 22.21 & 13.18 \\

CoSpaDi (grouped)                 &                             & 71.0 (68.8) & 58.5 (42.3) & 64.1 & 63.5 (67.4) & 35.5 (33.4) & 87.7 (91.3) & 35.8 & 24.0 & \textbf{55.0 (53.4)} & \textbf{15.43} & \textbf{6.51} \\ \hline

SVD-LLM                           & \multirow{4}{*}{0.4}        & 63.3 (62.9) & 42.9 (33.4) & 33.0 & 37.7 (41.9) & 24.7 (21.3) & 70.1 (77.0) & 32.3 & 23.0 & 40.9 (40.6) & 86.50 & 66.91 \\

CoSpaDi (per-layer)          &                                  & 66.0 (65.1) & 47.9 (36.9) & 46.4 & 44.4 (48.9) & 26.5 (22.7) & 76.7 (84.0) & 33.4 & 23.5 & 45.6 (45.1) & 50.46 & 21.50 \\\cdashline{3-13}
Basis Sharing                     &                             & 60.7 (60.3) & 41.5 (33.0) & 41.0 & 44.6 (49.1) & 26.5 (23.5) & 75.4 (82.8) & 30.1 & 23.2 & 42.9 (42.9) & 39.58 & 36.49 \\

CoSpaDi (grouped)                 &                             & 64.8 (63.4) & 48.0 (36.1) & 51.8 & 51.9 (56.7) & 29.1 (25.6) & 80.5 (88.2) & 32.7 & 23.0 & \textbf{47.7 (47.2)} & \textbf{25.29} & \textbf{14.99} \\ \hline
\end{tabular}
}
\caption{Performance comparison of CoSpaDi vs Basis Sharing~\citep{wang2025basis} and per-layer counterparts on LLaMA-2-7B under different CRs on various benchmarks. Best results are provided in \textbf{bold}.
Unnormalized accuracies are shown in parentheses for direct comparison with~\citep{wang2025basis}.}
\label{tab:main_results_group}\vspace{-.3cm}
\end{table*}

\subsection{Main Results}
\label{sec:main_results}

\subsubsection{Per-Layer Compression}
\label{sec:per_layer_results}

We first compress each projection matrix independently and compare CoSpaDi to SVD-LLM over a range of CRs.
Table~\ref{tab:main_results} reports results on LLaMA-3-8B and Qwen3-8B at CR 0.2--0.4, while additional results for smaller and larger models are provided in Appendix~\ref{apx:more_models}.
Across model families and compression budgets, CoSpaDi consistently improves both average benchmark accuracy and perplexity at matched CR. 
This suggests that the union-of-subspaces factorization preserves task-relevant directions more effectively than a single shared low-rank basis. 
We further report results on modern instruction-following and reasoning benchmarks in Appendix~\ref{apx:more_bench}, including IFEval~\citep{ifeval}, BBH~\citep{bbh}, MATH~\citep{math}, GPQA~\citep{gpqa}, MUSR~\citep{musr}, and MMLU-Pro~\citep{mmlupro}.

To position CoSpaDi beyond structured weight factorization, we also compare against recent structured pruning methods. 
Specifically, Table~\ref{tab:main_llama3_8b_train_free} includes ReplaceMe~\citep{shopkhoev2025replaceme} and LLM-Pruner~\citep{ma2023llmpruner}, which reduce parameters by removing or replacing structured components such as blocks or layers. 
These methods provide a complementary reference point under standard parameter-count compression, since their compressed models can still be evaluated within conventional dense inference pipelines.

Unstructured pruning methods involve additional deployment choices, such as sparse storage format, metadata representation, sparsity pattern, and kernel support. 
Since these choices can substantially affect the realized memory and latency at a fixed nominal sparsity level, we treat sparse pruning as a separate comparison rather than mixing storage conventions in the main table. 
For completeness, Appendix~\ref{apx:struct_pruning} reports a comparison with Wanda~\citep{sun2024wanda} in a 2:4 semi-structured setting, which provides a hardware-friendly sparse baseline.

\subsubsection{Cross-Layer Dictionary Sharing}
\label{sec:grouped_results}
Next, we evaluate grouped compression where a single dictionary is shared across multiple layers of the same projection type.
For a fair and controlled comparison, we adopt the \emph{exact} layer grouping and evaluation protocol of Basis Sharing~\citep{wang2025basis}, i.e., we share parameters across the same layer sets and report results under the same compression budgets.
Table~\ref{tab:main_results_group} reports results on LLaMA-2-7B across CR 0.2--0.4. We additionally included per-layer counterparts to show the benefits of grouping strategy and exploiting inter-layer redundancies.
To facilitate direct comparison with Basis Sharing, we report unnormalized accuracies in parentheses alongside the normalized results.

At all budgets, CoSpaDi substantially outperforms Basis Sharing as well as per-layer baselines, suggesting that cross-layer sharing is most effective when paired with sparse codes that allow each column (and each layer) to select its own subset of atoms.
We note that CoSpaDi does not rely on this particular grouping and could benefit from more adaptive sharing strategies (e.g., data-driven grouping or partially shared dictionaries), which we leave for future work.

\subsubsection{End-to-end Throughput}
\label{sec:e2e_throughput}

Beyond parameter reduction, practical compression depends on whether the induced computation can be executed efficiently.
We therefore evaluate a hardware-aware setting on LLaMA-3-8B, where only the wide MLP expansion projections (\texttt{up} and \texttt{gate}) are replaced by compressed modules.
This setting targets the projection types where CoSpaDi's dense-times-sparse structure is most favorable; ~\ref{apx:deployment} provides full details on the deployment setup, prompt lengths, and selective-compression protocol.
We denote variants that compress only the \texttt{up} and \texttt{gate} projections by the superscript $\mathrm{UG}$.

Table~\ref{tab:e2e_hw_speedup} summarizes the quality--throughput trade-off in synchronous generation.
At matched global compression ratios, CoSpaDi substantially outperforms SVD-LLM in both average accuracy and perplexity.
SVD-LLM remains faster in raw tokens/s under this implementation, but CoSpaDi maintains practical throughput while preserving much stronger model quality.
For reference, we include the dense LLaMA-3-8B baseline measured under the same deployment setup.

\begin{table}[t]
\centering
\resizebox{\columnwidth}{!}{%
\renewcommand{\arraystretch}{1.05}
\begin{tabular}{lcccccc}
\hline
Method & Global CR & UG CR & Avg.$\uparrow$ & Wiki$\downarrow$ & LAMB.$\downarrow$ & Tok/s$\uparrow$ \\
\hline
Dense baseline & -- & -- & 70.4 & 7.3 & 3.1 & 44.0 \\
\hline
SVD-LLM$^{\mathrm{UG}}$ & 0.22 & 0.40 & 53.5 & 56.2 & 8.5 & \textbf{53.3} \\
CoSpaDi$^{\mathrm{UG}}$ & 0.22 & 0.40 & \textbf{61.9} & \textbf{25.0} & \textbf{2.8} & 48.0 \\
\hline
SVD-LLM$^{\mathrm{UG}}$ & 0.27 & 0.50 & 43.8 & 213.0 & 54.5 & \textbf{52.5} \\
CoSpaDi$^{\mathrm{UG}}$ & 0.27 & 0.50 & \textbf{54.8} & \textbf{52.0} & \textbf{5.8} & 50.5 \\
\hline
\end{tabular}
}
\caption{Selective \texttt{up}/\texttt{gate} compression on LLaMA-3-8B. Throughput is averaged over prompt lengths $1$--$256$ in synchronous generation; higher is better.}
\label{tab:e2e_hw_speedup}\vspace{-0.3cm}
\end{table}

We also evaluate asynchronous serving in Table~\ref{tab:e2e_async_prompt1}, reporting prompt-length-1 throughput for the dense model, SVD-LLM$^{\mathrm{UG}}$, and CoSpaDi$^{\mathrm{UG}}$. 
Both compressed variants improve over the dense baseline at most concurrency levels, while CoSpaDi remains competitive with SVD-LLM and preserves substantially better quality. 
The full asynchronous breakdown is provided in Appendix~\ref{apx:deployment}.

\begin{table}[t]
\centering
\resizebox{\columnwidth}{!}{%
\renewcommand{\arraystretch}{1.05}
\begin{tabular}{lcccc}
\hline
Method & 1 user & 4 users & 8 users & 16 users \\
\hline
Dense baseline          & 77 & 290 & 589 & 1344 \\
SVD-LLM$^{\mathrm{UG}}$ & \textbf{99} & \textbf{386} & \textbf{798} & 1332 \\
CoSpaDi$^{\mathrm{UG}}$ & 96 & 376 & 758 & \textbf{1515} \\
\hline
\end{tabular}
}
\caption{Asynchronous serving throughput on LLaMA-3-8B for prompt length $1$. SVD-LLM$^{\mathrm{UG}}$ and CoSpaDi$^{\mathrm{UG}}$ use UG CR $=0.60$; higher is better.}
\label{tab:e2e_async_prompt1}\vspace{-0.3cm}
\end{table}

These results suggest that CoSpaDi's advantage comes from a stronger quality--compression trade-off that remains executable in an end-to-end setting.
Since CoSpaDi evaluates projections as $(\m X\m D_a)\m S$, its throughput is directly tied to sparse coefficient computation.
As sparse kernels and inference engines continue to improve~\citep{macko2025macko,chen2025elsa}, the measured throughput should be viewed as an initial deployment result rather than the systems limit of the proposed factorization.
\section{Conclusions}
\label{sec:conclusions}

CoSpaDi establishes activation-aware sparse dictionary learning as a new post-training compression paradigm beyond single-subspace low-rank approximation. 
Our current implementation intentionally uses a simple instantiation with fixed $\rho=k/s$, mostly uniform budgets, and a standard alternating solver. 
This leaves many complementary advances from low-rank compression such as dynamic allocation, improved calibration objectives, cross-layer sharing, and quantization-aware storage directly applicable to CoSpaDi without changing its core union-of-subspaces factorization.

From a systems perspective, CoSpaDi exposes a dense-times-sparse computation pattern whose practical efficiency depends on sparse runtimes and kernel support. 
Our deployment results in Appendix~\ref{apx:deployment} show measurable speedups over the dense baseline under hardware-aware choices, while ongoing progress in sparse kernels and inference engines suggests that these results are unlikely to be the final systems limit. 
Thus, CoSpaDi provides both a stronger compression model today and a flexible foundation for future advances in allocation, quantization, and sparse inference.

\section*{Limitations}
\label{sec:limitations}

\textbf{Solver efficiency and scalability.}
CoSpaDi relies on alternating minimization with OMP-based sparse coding and K-SVD-style dictionary updates. 
While this procedure is simple and effective, it is more expensive than closed-form SVD because it repeatedly alternates between sparse coding and per-atom dictionary refinement. 
Our runtime breakdown shows that the dictionary-update stage is the main bottleneck, reflecting the sequential nature of K-SVD.
This limitation is algorithmic rather than inherent to the CoSpaDi objective. 
Several established acceleration routes could reduce compression time, including joint dictionary updates such as MOD~\citep{engan1999mod}, efficient batched sparse coding such as Batch-OMP~\citep{rubinstein2008efficient}, online or stochastic dictionary learning~\citep{mairal2009online}, and approximate rank-one updates based on randomized SVD or power iterations~\citep{halko2011randomized,golub2013matrix}.
Our ablations already show that power-iteration updates provide a favorable accuracy-runtime trade-off.

\textbf{Budget allocation and hyperparameter selection.}
In the main experiments, we use a fixed ratio $\rho=k/s$ and uniform compression budgets to isolate the effect of the proposed union-of-subspaces factorization. 
However, prior low-rank work shows that layer-wise and type-wise sensitivity can substantially improve quality at a fixed global budget. 
Adapting such allocation strategies to CoSpaDi, for example by choosing $(k_\ell,s_\ell)$ per layer, projection type, or shared group is a natural extension. 
Similarly, more adaptive choices of $\rho$ may improve the trade-off between dictionary capacity and per-column sparsity.

\textbf{Grouped sharing design.}
Our grouped variant follows a simple sharing scheme aligned with prior cross-layer factor sharing. 
Its performance can depend on how layers are clustered and which projections share a dictionary. 
More adaptive grouping based on activation similarity, weight similarity, projection type, or learned sharing patterns could further improve the grouped setting.

Overall, these limitations reflect the direct instantiation studied in this work rather than fundamental restrictions of the sparse dictionary factorization itself.


\bibliography{custom,anthology-2}

\appendix

\section{Appendix}

\subsection{Data-Aware Low-Rank Weight Approximation}\label{apx:data_aware}
While the low-rank approximation of the weights $\m W$ has been extensively used for compression tasks, in practice is not well suited to LLMs and it can lead to a severe drop of their performance. Several recent works~\cite{chen2021drone,yuan2023asvd,wang2025svdllm} have suggested that instead of approximating the weights $\m W$  with a low-rank matrix, a more efficient strategy is to model the weight activations, $\m Z=\m X\m W$, as low-rank. Here, the matrix $\m X=\bbmtx \m x_1 &\ldots & \m x_N\ebmtx^\transp\in\R^{N\times d}$ holds in its rows the $d$-dimensional input vectors $\m x_n$ with $n=1\ldots, N$, which play the role of calibration data. Under this modeling framework, we can approximate the matrix weights $\m W$ as the minimizer of the following problem:
\bal
\begin{aligned}
\tilde{\m W}^\ast
&=
\argmin_{\tilde{\m W}}
\norm{\m X\m W-\m X\tilde{\m W}}{F}
\\[-0.25em]
&\quad
\text{s.t.}\quad
\rank\pr{\m X\tilde{\m W}}=r .
\end{aligned}
\label{eq:data_aware_loss}
\eal

Let us now consider $\m Y=\m X\m L^{-1}\in\R^{N\times d_1}$ to be a semi-orthogonal matrix (column-orthogonal matrix), that is $\m Y^\transp\m Y=\m I_{d_1}$, which is obtained by linearly transforming the matrix $\m X$ using a non-singular matrix $\m L\in\R^{d_1\times d_1}$. Here we assume that $N \ge d$ and the matrix $\m X$ is of full rank. We note that there are different ways we can achieve this column-orthogonalization of $\m X$. Among them we can employ the QR/SVD decomposition on $\m X$ and the Cholesky/Eigen-value decomposition on $\m X^\transp\m X$ to compute a proper linear transformation $\m L$. By using the representation $\m X=\m Y\m L$ we can rewrite the problem of Eq.~\eqref{eq:data_aware_loss} as:
\bal
\begin{aligned}
\tilde{\m W}^\ast
&=
\argmin_{\tilde{\m W}}
\norm{\m Y\m L\m W-\m Y\m L\tilde{\m W}}{F}
\\[-0.25em]
&\quad
\text{s.t.}\quad
\rank\pr{\m Y\m L\tilde{\m W}}=r .
\end{aligned}
\label{eq:data_aware_loss_v2}
\eal

To solve the above minimization problem we first note that due to the orthonormal columns of $\m Y$, it can be expressed in the equivalent form:
\bal
\begin{aligned}
\tilde{\m W}^\ast
&=
\argmin_{\tilde{\m W}}
\norm{\bm\Gamma-\m L\tilde{\m W}}{F}
\\[-0.25em]
&\quad
\text{s.t.}\quad
\rank\pr{\m L\tilde{\m W}}=r .
\end{aligned}
\label{eq:data_aware_loss_equiv}
\eal
where $\bm\Gamma=\m L\m W$. Next, we introduce the auxiliary matrix $\tilde{\bm\Gamma}=\m L\tilde{\m W}$ and the problem in Eq.~\eqref{eq:data_aware_loss_equiv} becomes: 
\bal
\begin{aligned}
\tilde{\bm\Gamma}^\ast
&=
\argmin_{\tilde{\bm\Gamma}}
\norm{\bm\Gamma-\tilde{\bm\Gamma}}{F}
\\[-0.25em]
&\quad
\text{s.t.}\quad
\rank\pr{\tilde{\bm\Gamma}}=r .
\end{aligned}
\label{eq:data_aware_loss_equiv_final}
\eal
which is the orthogonal projection of $\bm\Gamma$ to the space of $r$-rank matrices. Given that $\tilde{\bm\Gamma}^\ast=\m L\tilde{\m W}^\ast$ and $\m L$ is invertible, we can now recover $\tilde{\m W}^\ast=\m L^{-1}\tilde{\bm\Gamma}^\ast$.

To conclude, if $\bm\Gamma$ admits the singular value decomposition $\bm\Gamma=\m U\bm\Sigma\m V^\transp$, then the optimal $r$-rank approximation of $\m W$ that minimizes the loss in Eq.~\eqref{eq:data_aware_loss} can be written in the form:
\bal
\tilde{\m W}=\m B\m C = \underbrace{\m L^{-1}\m U_r}_{\m B}\underbrace{\bm\Sigma_r\m V_r^\transp}_{\m C}.
\eal

We note that in this case, unlike the direct weight low-rank approximation, the matrix $\m B=\m L^{-1}\m U_r$ does not correspond to a basis of a subspace of $\R^{d_1}$, since its columns are no longer orthonormal, that is $\m B^\transp\m B = \m U_r^\transp\pr{\m L\m L^\transp}^{-1}\m U_r\ne \m I$ .

\paragraph{Numerically robust Gram factorization.}
In practice, the empirical Gram matrix $\m G=\m X^\transp \m X$ may be ill-conditioned or only positive semidefinite. Following the robust scaling step used in SVD-LLM, we first attempt a Cholesky factorization. If it fails, we compute the smallest eigenvalue $\lambda_{\min}(\m G)$ and apply the diagonal shift
\bal
    \m G_{\epsilon}
    =
    \m G
    +
    \max\{0,-\lambda_{\min}(\m G)+\epsilon\}\m I ,
\label{eq:robust_cholesky_shift}
\eal
with a small $\epsilon>0$, after which Cholesky is applied to $\m G_{\epsilon}$. Up to the transpose convention of Cholesky, this gives $\m L^\transp\m L=\m G_{\epsilon}$. Thus, the transform is exact when $\m G$ is positive definite and otherwise corresponds to a numerically stabilized activation-space objective.
\subsection{Derivation of the Optimal Pair of Basis and Coefficient Matrices}\label{apx:PCA_proof}
Let us consider the weight matrix $\m W$ as a collection of $d_1$-dimensional vectors 
$\m W = \bbmtx \m w_1, &\ldots, \m w_{d_2}\ebmtx$, where $\bm w_j\in\R^{d_1}$, with $j=1,\ldots, d_2$. We seek to approximate each vector $\m w_j$ as a linear combination of basis vectors spanning a lower-dimensional subspace of $\R^{d_1}$. The optimal basis and coefficients can be found by minimizing the total approximation error:
\bal
\begin{aligned}
\mc{J}
&=
\suml_{j=1}^{d_2}
\norm{
    \m w_j-\suml_{i=1}^r c_{i,j}\bm b_i
}{2}^2
\\[-0.25em]
&\equiv
\norm{\m W-\m B\m C}{F}^2 .
\end{aligned}
\eal
subject to the orthogonality constraint $\m B^\transp \m B = \m I$, where $\m B = \bbmtx \bm b_1, \ldots, \bm b_r \ebmtx \in \R^{d_1 \times r}$ is the basis matrix, $\m C \in \R^{r \times d_2}$ is the coefficient matrix with entries $c_{i,j}$, and $\m I \in \R^{r \times r}$ is the identity matrix. 

To solve this problem, we first reformulate the objective $\mc{J}$ into an equivalent form:
\bal
\begin{aligned}
\mc{J}
&=
\trace\pr{\m W^\transp\m W}
-
2\trace\pr{\m W^\transp\m B\m C}
\\[-0.25em]
&\quad
+
\trace\pr{\m C^\transp\m B^\transp\m B\m C}.
\end{aligned}
\label{eq:objective}
\eal

Next, we consider the basis matrix $\m B$ as fixed and compute the gradient of the objective w.r.t the matrix coefficient $\m C$ as
\bal
\grad{\mc J}{\m C} = 2\m B^\transp\m B\m C - 2\m B^\transp\m W.
\label{eq:objective_grad}
\eal
Setting the gradient to zero and taking into account the constraint $\m B^\transp\m B=\m I$, we can recover the optimum matrix coefficient as: $\m C = \m B^\transp\m W$. Now, putting back $\m C$ in Eq.~\eqref{eq:objective} and using 
$\m B^\transp\m B=\m I$, we get:
\bal
\begin{aligned}
\mc{J}
&=
\trace\pr{\m W^\transp\m W}
-
2\trace\pr{\m W^\transp\m B\m B^\transp\m W}
\\[-0.25em]
&\quad
+
\trace\pr{
\m W^\transp\m B
\m B^\transp\m B
\m B^\transp\m W
}
\\[-0.25em]
&=
\trace\pr{\m W^\transp\m W}
-
\trace\pr{\m W^\transp\m B\m B^\transp\m W}
\\[-0.25em]
&=
\trace\pr{\m W^\transp\m W}
-
\trace\pr{\m B^\transp\m W\m W^\transp\m B}.
\end{aligned}
\eal

Based on this we can recover the optimum basis matrix $\m B$ as the maximizer of the constrained problem:
\bal
\begin{aligned}
\m B^\star
&=
\argmax_{\m B}
\trace\pr{\m B^\transp\m W\m W^\transp\m B}
\\[-0.25em]
&\quad
\text{s.t.}\quad
\m B^\transp\m B=\m I .
\end{aligned}
\label{eq:argmax}
\eal

The above maximization problem enjoys a closed-form solution~\cite{Fan1949}, which is fully defined by the eigenvalues of the matrix $\m P=\m W\m W^\transp$. Specifically, the matrix $\m P\in\R^{d_1\times d_1}$, which is symmetric and positive semi-definite, admits the eigenvalue decomposition $\m P=\m U\bm\Lambda\m U^\transp$, with $\m U\in \R^{d_1\times d_1}$ holding the eigenvectors of $\m P$ in its columns. Then, the maximizer of Eq.~\eqref{eq:argmax} is recovered as $\m B^\star=\m U_r$ where $\m U_r\in \R^{d_1\times r}$ is a reduced version of $\m U$ formed with the $r$ eigenvectors corresponding to the largest eigenvalues of $\m P$. A useful observation is that the eigenvectors of $\m P$ exactly match the left-singular vectors of $\m W\in\R^{d_1\times d_2}$. Indeed, if $\m W$ admits the singular value decomposition $\m W=\m U\bm\Sigma\m V^\transp$, then we have that: $\m P=\m W\m W^\transp=\m U\bm\Sigma^2\m U^\transp\equiv \m U\bm\Lambda\m U^\transp$, with $\bm\Lambda=\bm\Sigma^2$. Therefore, instead of performing the eigenvalue decomposition on $\m P$ we can recover $\m U$ and consequently $\m U_r$ by computing the SVD of $\m W$. 

Finally, we can compute the optimum coefficient matrix as:
\bal
\begin{aligned}
\m C^\star
&=
\pr{\m B^\star}^\transp \m W
=
\m U_r^\transp \m W
\\[-0.25em]
&=
\m U_r^\transp
\m U\bm\Sigma\m V^\transp
=
\bbmtx \m I_r & \m O \ebmtx
\bbmtx
\bm\Sigma_r \m V_r^\transp \\
\bm\Sigma_{d-r}\m V_{d-r}^\transp
\ebmtx
\\[-0.25em]
&=
\bm\Sigma_r\m V_r^\transp .
\end{aligned}
\label{eq:optim_matrix_coeff}
\eal
where $\bm V_r\in\R^{d_2\times r}$ is a reduced version of $\m V$ formed with the $r$ right singular vectors of $\m W$ that correspond to its top-$r$ singular values, which are kept in the diagonal matrix $\bm\Sigma_r\in\R^{r\times r}$. 
\subsection{Pseudo Algorithm of the Proposed Method}\label{apx:algo}
\textbf{Goal.} Given a weight matrix $\m W \in\mathbb{R}^{d_1\times d_2}$ and a small calibration set $\m X \in\mathbb{R}^{N\times d_1}$, compute an activation-aware sparse–dictionary factorization $\m W \approx \m {\widetilde{W}}=\m {D_a} \m S$ under a target compression ratio. The procedure consists of whitening the activation objective, alternating sparse coding and dictionary updates on the whitened weights, and a final de-whitening step.

\paragraph{(1) Calibration and whitening.}
Compute an invertible transform $\m L \in\mathbb{R}^{d_1\times d_1}$ (e.g., via QR/SVD of $\m X$ or Cholesky of $\m X^\top \m X$) such that $\m Y = \m X \m L^{-1}$ has orthonormal columns ($\m Y^\top \m Y = \m I$). Left-multiply $\m W$ to obtain the whitened weights $\m W_L = \m L \m W$. Whitening converts the data-aware loss $\|\m X \m W - \m X \m {\widehat W}\|_F^2$ into a standard Frobenius objective $\|\m W_L - \m D_L \m S\|_F^2$ that is amenable to dictionary learning.

\paragraph{(2) Initialization.}
Initialize the whitened dictionary $\m {D_L^{(0)}}\in\mathbb{R}^{d_1\times k}$ (e.g., random permutation of columns of $\m W_L$) and set $\m {S}^{(0)}=0$. The pair $(k,s)$ is set from the target compression ratio via Eq.~\ref{eq:cr_sd_effective} optionally using the fixed ratio $\rho=k/s$.

\paragraph{(3) Alternating minimization.}
Repeat for $t=1,\dots,T$:
\emph{(a) Sparse coding.} For each column $j$, solve
\bal
\m s_j^{(t)}
&\in
\argmin_{\|\m s\|_0\le s}
\big\|
(\m W_L)_{:,j}
-
\m D_L^{(t-1)}\m s
\big\|_2^2 .
\eal
using OMP (greedy selection with orthogonal residual updates) to enforce exactly $s$ nonzeros per column. \emph{(b) Dictionary update.} For each atom $i$, collect its support $\Omega_i=\{j:\, s_{i,j}^{(t)}\neq 0\}$ and form the residual on those columns:
\bal
\m R_i
&=
\m W_L[:,\Omega_i]
-
\sum_{\ell\neq i}
\m D_{L,\ell}^{(t-1)}
\m s_{\ell,\Omega_i}^{(t)} .
\eal
Update $(\m {D_{L,i}}^{(t)},\, \m {s}_{i,\Omega_i}^{(t)})$ by the best rank-1 approximation $\m R_i \approx \m u\, \m {\sigma}\, \m v^\top$ (set $\m D_{L,i}^{(t)}\!\leftarrow \m u$, $\m s_{i,\Omega_i}^{(t)}\!\leftarrow \m {\sigma} \m v^\top$). This preserves the current sparsity pattern while reducing the residual. Iterate atoms sequentially. Stop when the maximum iteration $T$ is reached or when the relative improvement falls below a tolerance.

\paragraph{(4) De-whitening and packing.}
Map the dictionary back to activation space via $ \m {D_a} = \m {L^{-1}} \m {D_L^{(T)}}$ and set $\widetilde{ \m W} = \m D_a \m S^{(T)}$. For storage, keep $\m D_a$ and the $s d_2$ nonzero entries of $\m S$ in \texttt{bf16} along with a packed binary mask $\m M\in\{0,1\}^{k\times d_2}$ for locations (one bit per entry; $\tfrac{k d_2}{16}$ words). This yields the compression ratio in Appendix~\ref{apx:cr} and Eq.~\ref{eq:cr_sd_effective}.

\paragraph{(5) Inference.}
At runtime, apply $\widetilde{\m W}$ as $\mathrm{matmul}(\m x, \m D_a \m S)$ with sparse–dense kernels. Reuse inner products $\langle \m x, \m {D_{a,:,i}}\rangle$ across columns to achieve the complexity in Appendix~\ref{apx:inference_complexity}; the number of active atoms controls the practical speedup.

\begin{algorithm*}[!h]
\caption{Pseudocode for CoSpaDi. The alternating optimization consists
of (a) sparse coding and (b) sequential dictionary updates.}
\label{alg:cospadi}
\Input{
    $\m W \in \R^{d_1 \times d_2}$: weight matrix to compress \\
    $\m X \in \R^{N \times d_1}$: calibration input data ($N$ samples) \\
    $k$: dictionary size (number of atoms, $k \geq s$) \\
    $s$: sparsity level (max nonzeros per column in $\m S$) \\
    $T$: number of K-SVD iterations
}
\Output{
    $\m D_a \in \R^{d_1 \times k}$: activation-aware dictionary \\
    $\m S \in \R^{k \times d_2}$: sparse coefficient matrix \\
    $\tilde{\m W} = \m D_a \m S$: compressed weight matrix
}

Compute $\m L \in \R^{d_1 \times d_1}$ such that $\m Y = \m X \m L^{-1}$ satisfies $\m Y^\transp \m Y = \m I_{d_1}$\;
\Comment{e.g., via QR: $\m X = \m Q \m R \Rightarrow \m L = \m R$}
\Comment{e.g., via Cholesky: $\m X^\transp\m X = \m C^\transp \m C \Rightarrow \m L = \m C$}

$\m W_L \gets \m L \m W$\;
Initialize $\m D_L^{(0)} \in \R^{d_1 \times k}$ from randomly selected columns of $\m W_L$\;
Initialize $\m S^{(0)} \in \R^{k \times d_2}$ as zero matrix\;

\For{$t = 1$ \KwTo $T$}{
    \For{$j = 1$ \KwTo $d_2$}{
        $\m s_j^{(t)} \gets \argmin_{\norm{\m s}{0} \leq s} \norm{ \m W_{L,j} - \m D_L^{(t-1)} \m s }{2}^2$\;
        \Comment{Solve via batched OMP}
    }

    \For{$i = 1$ \KwTo $k$}{
        $\Omega_i \gets \cbr{ j \,\vert\, s_{i,j}^{(t)} \neq 0 }$\;
        \If{$\Omega_i \neq \emptyset$}{
            $\m R_i \gets \m W_L[:, \Omega_i] - \sum_{l \neq i} \m d_{L,l}^{(t-1)} \m s_{l,\Omega_i}^{(t)}$\;
            $[\m u, \sigma, \m v] \gets \text{rank-1 SVD of } \m R_i$\;
            $\m d_{L,i}^{(t)} \gets \m u$\;
            $\m s_{i,\Omega_i}^{(t)} \gets \sigma \cdot \m v^\transp$\;
        }
    }
}

$\m D_a \gets \m L^{-1} \m D_L^{(T)}$\;

$\tilde{\m W} \gets \m D_a \m S^{(T)}$\;

\Return $\m D_a$, $\m S^{(T)}$, $\tilde{\m W}$\;
\end{algorithm*}
\subsection{Optimization Behavior and Convergence}
\label{apx:convergence}

The CoSpaDi objective in Eq.~\ref{eq:DicLearn_data_aware_final} is non-convex because of the bilinear factorization and the column-wise sparsity constraints. Therefore, we do not claim convergence to a global optimum. Nevertheless, the alternating structure admits a standard monotonicity result when the two block subproblems are solved exactly.

Let
\begin{equation}
    F(D_L,S)
    =
    \|W_L-D_LS\|_F^2,
\end{equation}
and denote the feasible set of sparse coefficient matrices by
\begin{equation}
    \mathcal{S}_s
    =
    \left\{
        S:\|s_j\|_0\leq s,\ \forall j
    \right\}.
\end{equation}
Consider the idealized alternating updates
\begin{equation}
    S^{(t+1)}
    \in
    \arg\min_{S\in\mathcal{S}_s}
    F(D_L^{(t)},S),
    \label{eq:exact_sparse_update}
\end{equation}
followed by
\begin{equation}
    D_L^{(t+1)}
    \in
    \arg\min_{D}
    F(D,S^{(t+1)}).
    \label{eq:exact_dict_update}
\end{equation}
By optimality of the sparse-coding update,
\begin{equation}
    F(D_L^{(t)},S^{(t+1)})
    \leq
    F(D_L^{(t)},S^{(t)}),
\end{equation}
and, similarly, the dictionary update gives
\begin{equation}
    F(D_L^{(t+1)},S^{(t+1)})
    \leq
    F(D_L^{(t)},S^{(t+1)}).
\end{equation}
Combining the two inequalities yields
\begin{equation}
    F(D_L^{(t+1)},S^{(t+1)})
    \leq
    F(D_L^{(t)},S^{(t)}).
    \label{eq:cospa_monotonicity}
\end{equation}
Hence, the reconstruction objective is monotonically non-increasing under exact block minimization. Since $F(D_L,S)\geq0$, the sequence of objective values converges to a finite limit.

This result concerns convergence of the \emph{objective values}, rather than convergence to the globally optimal dictionary and sparse codes. The sparsity-constrained problem remains non-convex, and different initializations may lead to different solutions. More generally, connections can be made to classical block-coordinate descent results, where stronger statements about limit points require additional assumptions on the block subproblems and iterates~\citep{tseng2001convergence}. We therefore do not rely on a global or unique-stationary-point guarantee.

\paragraph{Practical OMP/K-SVD updates.}
The implementation used in our experiments follows the standard practical dictionary-learning procedure rather than solving both block problems globally. In particular, OMP provides a greedy approximation to the sparse-coding problem in Eq.~\ref{eq:sparse_coding}~\citep{tropp2007omp,rubinstein2008efficient}. Consequently, the exact-block monotonicity argument above does not by itself guarantee that every OMP sparse-coding update decreases the objective relative to the preceding iterate.

For the dictionary step, standard K-SVD updates one atom and its active coefficients at a time. With the support of an atom fixed, the exact rank-one SVD update minimizes the corresponding residual and therefore cannot increase the reconstruction error during that atom update~\citep{aharon2006ksvd}. Our main implementation replaces the exact rank-one SVD with a small number of power iterations for efficiency; this is again an approximate update and therefore does not provide an unconditional exact-descent guarantee.

Thus, Eq.~\eqref{eq:cospa_monotonicity} characterizes the underlying exact alternating-minimization scheme, while the OMP/K-SVD procedure used in practice inherits the standard approximate behavior of classical dictionary learning. If strict monotonicity is required, it can be enforced by accepting a sparse-code or dictionary update only when it does not increase the reconstruction residual. Empirically, we observe stable optimization with the adopted procedure; the sensitivity to the number of alternating and power iterations is reported in Appendix~\ref{apx:iter_ablation}.

\subsection{Grouped CoSpaDi details}\label{apx:grouped}

The exact shared-dictionary objective for a group $\mathcal{G}$ is
\begin{equation}
\begin{aligned}
    \min_{\m D,\{\m S_\ell\}_{\ell\in\mathcal{G}}}
    &\sum_{\ell\in\mathcal{G}}
    \|\m X_\ell\m W_\ell
    -
    \m X_\ell\m D\m S_\ell\|_F^2
    \\
    \text{s.t.}\quad
    &\|\m s_{\ell,j}\|_0\le s,
    \quad \forall \ell\in\mathcal{G},\ \forall j .
\end{aligned}
\label{eq:appendix_grouped_exact}
\end{equation}
Writing $\m G_\ell=\m X_\ell^\top\m X_\ell$, this is equivalently
\begin{equation}
    \sum_{\ell\in\mathcal{G}}
    \mathrm{tr}
    \left[
        (\m W_\ell-\m D\m S_\ell)^\top
        \m G_\ell
        (\m W_\ell-\m D\m S_\ell)
    \right].
    \label{eq:appendix_grouped_gram_form}
\end{equation}
Unlike the per-layer case, different layers induce different Gram metrics. 
With fixed sparse codes, the exact dictionary update satisfies
\begin{equation}
    \sum_{\ell\in\mathcal{G}}
    \m G_\ell\m D(\m S_\ell\m S_\ell^\top)
    =
    \sum_{\ell\in\mathcal{G}}
    \m G_\ell\m W_\ell\m S_\ell^\top ,
    \label{eq:appendix_grouped_normal}
\end{equation}
or, after vectorization,
\begin{equation}
\begin{aligned}
&\left[
\sum_{\ell\in\mathcal{G}}
(\m S_\ell\m S_\ell^\top)^\top
\otimes \m G_\ell
\right]
\mathrm{vec}(\m D)
\\[-0.25em]
&\qquad =
\mathrm{vec}
\left[
\sum_{\ell\in\mathcal{G}}
\m G_\ell\m W_\ell\m S_\ell^\top
\right].
\end{aligned}
\label{eq:appendix_grouped_vectorized}
\end{equation}
Solving this system is expensive for LLM-scale projections. 
We therefore use the shared-metric approximation from Section~\ref{sec:optimization_grouping}, where
\begin{equation}
    \bar{\m G}_{\mathcal{G}}
    =
    \frac{1}{|\mathcal{G}|}
    \sum_{\ell\in\mathcal{G}}
    \m G_\ell
    =
    \bar{\m L}^\top\bar{\m L}.
\end{equation}
This replaces Eq.~\eqref{eq:appendix_grouped_exact} by
\begin{equation}
\begin{aligned}
    \min_{\m D,\{\m S_\ell\}}
    &\sum_{\ell\in\mathcal{G}}
    \|\bar{\m L}(\m W_\ell-\m D\m S_\ell)\|_F^2
    \\
    \text{s.t.}\quad
    &\|\m s_{\ell,j}\|_0\le s,
    \quad \forall \ell\in\mathcal{G},\ \forall j .
\end{aligned}
\label{eq:appendix_grouped_shared_metric}
\end{equation}
Defining $\m D_L=\bar{\m L}\m D$ and concatenating transformed weights yields
\begin{equation}
\begin{aligned}
    \min_{\m D_L,\m S_{\mathcal{G}}}
    &\|\m W_{\mathcal{G},L}
    -
    \m D_L\m S_{\mathcal{G}}\|_F^2
    \\
    \text{s.t.}\quad
    &\|\m s_{\ell,j}\|_0\le s,
    \quad \forall \ell,j .
\end{aligned}
\label{eq:appendix_grouped_standard}
\end{equation}
which has the same form as the per-layer CoSpaDi objective and is solved with the same OMP/K-SVD alternating procedure. 
After optimization, $\m D_a=\bar{\m L}^{-1}\m D_L$, and each layer is reconstructed as $\tilde{\m W}_\ell=\m D_a\m S_\ell$.
\subsection{Derivation of the CoSpaDi Compression Ratio}\label{apx:cr}

We derive the expression for the compression ratio $\gamma^{\mathrm{SD}}$ of our sparse–dictionary (SD) parameterization. Let $\m W\!\in\!\mathbb{R}^{d_1\times d_2}$ be factorized as
\bal
\m W \;\approx\; \m D \m S, \qquad
\m D\in\mathbb{R}^{d_1\times k},\ \ \m S\in\mathbb{R}^{k\times d_2},
\eal
where each column of $\m S$ has exactly $s$ nonzeros (\emph{column-$s$-sparse}). Throughout, we store real values in \texttt{bfloat16} (16\,bits) as is common in modern LLMs.

\paragraph{Dense baseline.}
A dense $\m W$ requires $d_1 d_2$ \texttt{bf16} values.

\paragraph{Dictionary term.}
The dictionary $\m D$ stores $d_1 k$ \texttt{bf16} values.

\paragraph{Sparse codes.}
Naively, $\m S$ would need $k d_2$ values. Since $\m S$ is column-$s$-sparse, only $s d_2$ values are stored. For locations, one option is COO: per nonzero we keep a row index and (redundantly) the column index. Because sparsity is fixed per column, column indices can be omitted; keeping only row indices yields $s d_2$ indices. With 16-bit indices, the total becomes $s d_2$ values $+$ $s d_2$ indices $= 2 s d_2$ 16-bit words. For typical $\rho\!\coloneqq\!k/s=2$, this equals $k d_2$ words—offering no savings over dense $\m S$ storage.

Instead, we use a \emph{bit mask} $\m M\!\in\!\{0,1\}^{k\times d_2}$ to mark nonzero positions. This requires $k d_2$ bits, i.e., $\nicefrac{k d_2}{16}$ 16-bit words after packing. We then store $s d_2$ \texttt{bf16} values for the nonzeros and the packed mask for their positions.

\paragraph{Total and ratio.}
The SD parameterization thus stores
\bal
\underbrace{d_1 k}_{\text{dictionary}} \;+\;
\underbrace{s d_2}_{\text{values}} \;+\;
\underbrace{\tfrac{k d_2}{16}}_{\text{mask}}
\eal
16-bit words. Relative to the dense baseline $d_1 d_2$, the resulting compression ratio is
\bal
\gamma^{\mathrm{SD}}\coloneqq 1-\tfrac{\;\overbrace{d_1k}^{\text{dict. (bf16)}}+\overbrace{sd_2}^{\text{codes (bf16)}}+\overbrace{(kd_2)/16}^{\text{mask (1 bit/entry)}}\;}{d_1d_2}.
\eal
\paragraph{Mask compensation by coefficient truncation.}
In our main experiments we use $\rho\coloneqq k/s=2$. 
Therefore, the packed binary mask requires
\bal
kd_2 = 2sd_2
\eal
bits. 
At the same time, truncating two mantissa bits from each of the $sd_2$ stored coefficient values saves exactly
\bal
2sd_2
\eal
bits. 
Thus, for $\rho=2$, the storage cost of the binary mask is exactly compensated by the two-bit coefficient truncation:
\bal
&\underbrace{14sd_2}_{\text{trunc. coef. values}}
+
\underbrace{kd_2}_{\text{mask}}
\\[-0.25em]
&\qquad =
14sd_2+2sd_2
=
16sd_2
\quad \text{bits}.
\eal
Equivalently, the truncated coefficient values together with their packed mask occupy the same storage as $sd_2$ bf16 values. 
This gives the effective compression ratio used in our main experiments:
\bal
\hat{\gamma}^{\mathrm{SD}}
=
1-
\frac{
d_1k+sd_2
}{
d_1d_2
}.
\label{eq:appendix_sd_cr_effective}
\eal
This accounting relies on both assumptions: $\rho=2$ and two-bit coefficient truncation. 
Without truncation, the mask contributes an additional $(kd_2)/16$ bf16 words; for $\rho\neq 2$, truncating two bits no longer exactly compensates the mask term.
This matches the expression used in the main text and makes explicit the dependence on the two design knobs $(k,s)$.
\subsection{Low-rank and CoSpaDi Inference Complexity} \label{apx:inference_complexity}

Here we derive the multiplication complexity for the original weight, SVD-compressed weight, and dictionary-learning (k-SVD) compression. We count \underline{multiplications only} (additions are of the same order). Let $\m W \in \R^{d_1 \times d_2}$ be a projection matrix in some layer and $\m X \in \R^{N \times d_1}$ be an input feature map; then a dense product $\m Y = \m X \m W$ costs
\bal
O_{baseline} = N d_1 d_2.
\eal
For low-rank, in particular, SVD compression with rank $r$ the projection matrix is approximated with two matrices $\m W \approx \m U \m V$ with $\m U \in \R^{d_1 \times r}$ and $\m V \in \R^{r \times d_2}$, resulting in the following complexity:
\bal
O_{LR} = N d_1 r + N r d_2 = N r (d_1 + d_2).
\eal
Sparse dictionary (SD) learning similarly represents $\m W \approx \m D \m S$ with dictionary $\m D \in \R^{d_1 \times k}$ of $k$ atoms and sparse coefficient matrix $\m S \in \R^{k \times d_2}$. Omitting sparsity of $\m S$ will result in:
\bal
O_{SD,\mathrm{dense}}
&=
N d_1 k + N k d_2 \\
&=
N k (d_1+d_2).
\eal
Taking into account that each column $\m s_j$ of $\m S$ is $s$-sparse, the $(i,j)$ element of $\m Y = \m X \m D \m S$ is
\bal
\begin{aligned}
y_{i,j}
&=
\sum_{k=1}^{K}
\m S_{k,j}
\left\langle \m X_{i,:}, \m D_{:,k} \right\rangle
\\
&=
\sum_{k\in\mathcal{S}_j}
\m S_{k,j}
\left\langle \m X_{i,:}, \m D_{:,k} \right\rangle .
\end{aligned}
\eal
where $\mathcal{S}_j=\mathrm{supp}(s_j)$ and $|\mathcal{S}_j| = s$. The overall sparse complexity depends on whether the inner products $\langle \m X_{i,:}, \m D_{:,k}\rangle$ are reused across columns. With the most efficient way with reuse letting $\m U = \bigcup_{j = 1} ^ {d_2} \mathcal{S}_j$ and $K_{active}=|U|$ we have:
\bal
\begin{alignedat}{2}
O_{SD,\mathrm{sparse\text{-}reuse}}
&= N d_1 K_{\mathrm{active}} + N s d_2,\\
s
&\le K_{\mathrm{active}} \le \min(K,sd_2).
\end{alignedat}
\label{eq:sd_sparse_reuse}
\eal

Using the effective storage accounting from Appendix~\ref{apx:cr}, where $\rho=2$ and two-bit coefficient truncation compensates the packed mask cost, we have
\bal
\hat{\gamma}^{\mathrm{SD}}
=
1-
\frac{d_1k+sd_2}{d_1d_2}.
\eal
The rank for the low-rank decomposition could be estimated from the compression ratio with the following equation:
\bal
r=\frac{(1 - \gamma^{LR}) d_1 d_2}{d_1 + d_2}
\eal
For sparse dictionary based method we defined $\rho=k/s$ and, thus we can estimate both $k$ and $s$ in the following way:
\bal
k=\frac{(1-\hat{\gamma}^{\mathrm{SD}})\,d_1d_2}{\,d_1+\tfrac{d_2}{\rho}\,},\qquad
s=\tfrac{k}{\rho}.
\label{eq:k_s}
\eal

Under matched effective storage, the sparse computation can be related to the low-rank multiplication count using the upper bound $K_{\mathrm{active}}\le k$. From Eq.~\eqref{eq:sd_sparse_reuse}, we have
\bal
\begin{aligned}
O_{SD,\mathrm{sparse\text{-}reuse}
}
&=
N d_1 K_{\mathrm{active}} + Nsd_2 \\
&\le
N d_1 k + Nsd_2 \\
&=
N\left(d_1k+d_2\frac{k}{\rho}\right) \\
&=
Nk\left(d_1+\frac{d_2}{\rho}\right).
\end{aligned}
\eal
Using the dictionary size implied by the effective storage in Eq. ~\eqref{eq:k_s}, 
we obtain
\bal
O_{SD,\mathrm{sparse\text{-}reuse}}
\le
N(1-\hat{\gamma}^{SD})d_1d_2.
\eal
For low-rank compression, Eq.~\eqref{eq:cr_lr} gives
\bal
O_{LR}
=
N(1-\gamma^{LR})d_1d_2.
\eal
Therefore, when the two methods are compared at the same effective compression ratio,
$\gamma^{LR}=\hat{\gamma}^{SD}=\gamma$, we have
\bal
O_{SD,\mathrm{sparse\text{-}reuse}}
\le
O_{LR}
=
N(1-\gamma)d_1d_2.
\eal
Equality in the upper-bound comparison is attained when
$K_{\mathrm{active}}=k$, i.e., when every dictionary atom is used by at least one coefficient column. If only a subset of the dictionary atoms is active, $K_{\mathrm{active}}<k$, the nominal sparse multiplication count is correspondingly smaller.

This comparison should be interpreted as a storage-aligned upper bound on the leading multiplication count rather than as a latency prediction. Actual wall-clock performance additionally depends on support overlap, indexing and mask-handling overhead, memory traffic, batching, and sparse-kernel efficiency. For this reason, we report measured end-to-end deployment results separately in Appendix~\ref{apx:deployment}, where the runtime setup, selective \texttt{up}/\texttt{gate} compression protocol, and throughput measurements are described explicitly.
\subsection{Relationship to Recent SVD-Based Extensions}
\label{apx:orthogonal_svd_extensions}

Our main comparison with SVD-LLM~\citep{wang2025svdllm} focuses on the representation used for matrix compression. SVD-based methods approximate each projection matrix using a single low-dimensional subspace, whereas CoSpaDi replaces this representation with a sparse dictionary and a union of subspaces. More recent methods, including SVD-LLM V2~\citep{wang2025svdllmv2} and Dobi-SVD~\citep{qinsi2025dobisvd}, extend the SVD-based compression pipeline along additional dimensions, such as heterogeneous rank allocation, loss-aware truncation, differentiable rank selection, and low-bit remapping. In this section, we discuss these differences and provide additional empirical comparisons.

\paragraph{SVD-LLM V2.}
SVD-LLM V2 extends SVD-LLM mainly through two mechanisms: \textit{non-uniform compression-ratio} allocation based on the estimated truncation loss and an improved \textit{loss-aware truncation} procedure. These mechanisms improve how the rank budget is distributed and how the low-rank approximation is computed, while retaining the underlying rank-constrained single-subspace representation.

The allocation mechanism is complementary to CoSpaDi. An analogous strategy could allocate layer-wise or group-wise parameters $(k_\ell,s_\ell)$ under a global storage budget instead of using the uniform allocation adopted in our main experiments. Consequently, a comparison with dynamically allocated SVD-LLM V2 reflects both the choice of factorization family and the effect of heterogeneous budget allocation. Developing such an allocation strategy for CoSpaDi is a natural direction for future work.

We additionally examined the numerical motivation for the revised SVD-LLM V2 truncation procedure under our calibration setting. Across 576 activation Gram matrices collected in our experiments, direct Cholesky factorization failed for only 3 matrices ($0.52\%$), all of which were handled by the stabilized factorization used in our implementation. Thus, in our experimental setting, heterogeneous allocation represents a particularly relevant difference between SVD-LLM and SVD-LLM V2.

For additional empirical context, we attempted to reproduce SVD-LLM V2 on LLaMA-7B. At the time of our experiments, the public SVD-LLM repository did not provide a dedicated ready-to-run SVD-LLM V2 pipeline, and several implementation conventions required interpretation. Our reproduction therefore did not exactly recover the values reported in the SVD-LLM V2 paper. We report both the published values and our reproduced results in Table~\ref{tab:svdllmv2_reference} for transparency.

\begin{table}[h]
\resizebox{0.45 \textwidth}{!}{%
\renewcommand{\arraystretch}{1.05}
\begin{tabular}{lcc}
\hline
\multirow{2}{*}{\textbf{Method}} & \multicolumn{2}{c}{\textbf{Perplexity $\downarrow$}} \\
                                 & \textbf{WikiText} & \textbf{C4} \\
\hline
\textbf{LLaMA-7B}              & 5.67 & 7.33  \\
\hline \hline
SVD-LLM V2, reported        & 7.12 & 10.47 \\
SVD-LLM V2, reproduced      & 8.09 & 16.29 \\
CoSpaDi                      & 8.12 & 10.26 \\
\hline
\end{tabular}
}
\caption{
Comparison with SVD-LLM V2 on LLaMA-7B at CR $=0.2$.
``Reported'' denotes the values reported in the SVD-LLM V2 paper, while
``reproduced'' denotes our implementation. Lower perplexity is better.
}
\label{tab:svdllmv2_reference}
\end{table}

The reproduced SVD-LLM V2 result is close to CoSpaDi on WikiText-2 but substantially worse on C4, while the paper-reported SVD-LLM V2 result performs better on WikiText-2 and similarly on C4. Because our reproduction does not exactly recover the published values and SVD-LLM V2 additionally changes the allocation policy, we include these results as complementary empirical context rather than using them to draw conclusions about the factorization family alone.

\begin{table*}[h]
\centering
\resizebox{0.8\textwidth}{!}{%
\renewcommand{\arraystretch}{1.05}
\begin{tabular}{lccccc}
\hline
Method
& Target CR
& Remapping
& WikiText-2 $\downarrow$
& C4 $\downarrow$
& Avg. Acc. $\uparrow$ \\ \hline

LLaMA-7B
& -- & --
& 5.67 & 7.33 & 0.550 \\ \hline \hline

Dobi-SVD$^\ast$, reported
& 0.2 & No
& 8.54 & 10.01 & 0.456 \\

Dobi-SVD, reported
& 0.2 & Yes
& 6.08 & 7.83 & 0.507 \\

Dobi-SVD, reproduced
& 0.2 & Yes
& 6.08 & 8.49 & 0.533 \\

CoSpaDi
& 0.2 & No
& 8.12 & 10.26 & 0.508 \\ \hdashline

Dobi-SVD$^\ast$, reported
& 0.4 & No
& 13.54 & 23.54 & 0.380 \\

Dobi-SVD, reported
& 0.4 & Yes
& 8.12 & 12.63 & 0.470 \\

Dobi-SVD, reproduced
& 0.4 & Yes
& 7.68 & 12.84 & 0.487 \\

CoSpaDi
& 0.4 & No
& 19.77 & 18.29 & 0.438 \\ \hline
\end{tabular}
}
\caption{Comparison with Dobi-SVD on LLaMA-7B.
Dobi-SVD$^\ast$ denotes Dobi-SVD without remapping, as reported in the original paper.
``Reproduced'' denotes our evaluation of the released remapped checkpoints.
Average accuracy is computed over OpenBookQA, ARC-easy, Winogrande, HellaSwag,
ARC-challenge, PIQA, and MathQA.
Dobi-SVD remapping changes the relation between factorization rank and final storage and is therefore indicated separately rather than expressed as a separate factorization compression ratio. For a square projection, the same remapped storage budget can correspond
to a negative conventional factorization compression ratio; see
Eq.~\eqref{eq:dobi_square_example}.} 
\label{tab:dobi_reference}
\end{table*}

\paragraph{Dobi-SVD.}
Dobi-SVD extends low-rank compression along two additional axes. First, it optimizes layer-wise ranks through a differentiable objective using \textit{backpropagation}. This differs from the training-free setting considered by CoSpaDi, where calibration samples are used to construct activation statistics but no gradient-based optimization of the compressed model is performed. Nevertheless, differentiable or sensitivity-based allocation is conceptually complementary to CoSpaDi and could similarly be used to allocate $(k_\ell,s_\ell)$ across layers.

Second, the strongest Dobi-SVD configurations combine low-rank
factorization with a low-bit remapping mechanism. This remapping changes
the relation between factorization rank and final storage relative to a
conventional two-factor SVD representation. Moreover, the corresponding
relation depends on the dimensions of each projection and on the
layer-wise rank allocation. We therefore do not assign a single separate
factorization compression ratio to the remapped LLaMA-7B model, and
instead compare the methods at their reported final storage budgets.

For intuition, consider a square projection
$W\in\mathbb{R}^{d\times d}$. Under Dobi-SVD remapping, a rank-$k$
representation has an effective retained-storage ratio $k/d$, whereas a
conventional two-factor SVD representation of the same rank requires
$2kd$ values and therefore has retained ratio $2k/d$. Denoting the final
remapped compression ratio by $\gamma_{\mathrm{target}}$, we have
$k/d=1-\gamma_{\mathrm{target}}$, and hence the corresponding
factorization-only compression ratio for this square-matrix example is
\begin{equation}
    \gamma_{\mathrm{fact}}
    =
    1-2(1-\gamma_{\mathrm{target}})
    =
    2\gamma_{\mathrm{target}}-1.
    \label{eq:dobi_square_example}
\end{equation}
Thus, for example, target compression ratios of $0.2$ and $0.4$
correspond to factorization-only compression ratios of $-0.6$ and
$-0.2$, respectively, for a square projection. A negative value means
that the conventional two-factor representation at the same rank would
contain more stored values than the original dense matrix before
remapping. This example illustrates how Dobi-SVD can use remapping to
support substantially larger ranks at a fixed final storage budget.
For non-square projections, however, the corresponding relation depends
on the matrix aspect ratio, which is why we avoid assigning a single
factorization compression ratio to the complete LLaMA-7B model.

For additional empirical comparison, we evaluated the released Dobi-SVD LLaMA-7B checkpoints. The official evaluation path produced numerical overflow for WikiText-2 perplexity in our environment, and we therefore evaluated the released checkpoints using the same evaluation pipeline employed for SVD-LLM.  We express Dobi-SVD's reported storage budgets using our compression-ratio convention, where CR denotes the fraction of storage removed relative to the dense model. Table~\ref{tab:dobi_reference} reports both the values from the Dobi-SVD paper and our reproduced results.

The results show that Dobi-SVD's remapping substantially improves the quality of its compressed models relative to its factorization-only variant. At the same time, the remapped results combine low-rank factorization with an additional storage mechanism, whereas the CoSpaDi rows correspond to the sparse-dictionary storage convention used throughout our main experiments. Since CoSpaDi can likewise be combined with post-training coefficient quantization, low-bit storage and remapping constitute an additional compression axis that is complementary to the sparse-dictionary representation studied here.

\paragraph{Summary.}
SVD-LLM V2 and Dobi-SVD extend SVD-based compression along dimensions that go beyond the underlying low-rank representation. SVD-LLM V2 introduces heterogeneous allocation and an improved loss-aware truncation procedure, while Dobi-SVD combines differentiable rank allocation with low-bit remapping. The comparisons above provide empirical context for CoSpaDi relative to these recent methods while making the differences in allocation and storage mechanisms explicit. These mechanisms are largely complementary to the union-of-subspaces representation introduced by CoSpaDi and could potentially be incorporated into future extensions of the method.
\subsection{Implementation details and reproducibility}\label{apx:impl_details}

This appendix consolidates the main implementation details used in the experiments.
Unless otherwise specified, CoSpaDi uses K-SVD with power-iteration-based rank-1 atom updates, with $T=60$ alternating iterations and 8 power iterations per update.
Sparse coding is performed with batched OMP using a mini-batch size of 8192.
The dictionary is initialized from a random permutation of columns of the original projection matrix.
All experiments use a fixed random seed of 42.

Calibration is performed using 256 sequences of length 1024 sampled from RefinedWeb.
Unless stated otherwise, the compressed models use bfloat16 dictionaries and truncated 14-bit coefficients, and the reported compression ratios follow the storage convention described in Section~\ref{sec:storage_complexity} and~\ref{apx:cr}. All hyperparameters are listed in Table~\ref{tab:hyperparams}

\begin{table}[t]
\centering
\resizebox{0.5\textwidth}{!}{%
\renewcommand{\arraystretch}{1.05}
\begin{tabular}{ll}
\hline
Hyperparameter & Value \\
\hline
Random seed & 42 \\
K-SVD iterations $T$ & 60 \\
Ratio $\rho = k/s$ & 2.0 \\
Dictionary update solver & power-iteration K-SVD \\
Power iterations & 8 \\
OMP batch size & 8192 \\
Target modules & linear layers in transformer blocks \\
Dictionary size $k$ & determined by CR and $\rho$ via Eq.~\eqref{eq:cr_sd_effective} \\
Sparsity $s$ & determined by CR and $\rho$ via Eq.~\eqref{eq:cr_sd_effective} \\
Dictionary initialization & random columns of $\m W_L$ \\
Calibration dataset & RefinedWeb \\
Number of calibration samples & 256 \\
Sequence length & 1024 \\
Storage convention for reported CR & Eq.~\eqref{eq:cr_sd_effective} unless stated otherwise \\
\hline
\end{tabular}
}
\caption{Main hyperparameters used in the experiments unless otherwise specified. The dictionary size $k$ and sparsity level $s$ are determined from the target compression ratio and the ratio $\rho = k/s$ via Eq.~\eqref{eq:cr_sd_effective}.}
\label{tab:hyperparams}
\end{table}
\subsection{Sensitivity to Calibration Data}
\label{apx:calibration_sensitivity}

CoSpaDi uses a small calibration set to construct the activation statistics that define the transformed dictionary-learning objective. In the main experiments, we use 256 sequences of length 1024 sampled from RefinedWeb. To examine whether the observed gains depend on this particular calibration setup, we study sensitivity along two axes: the calibration domain and the number of calibration samples. In all experiments below, we compare CoSpaDi and SVD-LLM using the same calibration data and a fixed compression ratio of $\mathrm{CR}=0.2$ on LLaMA-3.2-1B.

\paragraph{Calibration domain.}
We first vary the source of the calibration sequences while keeping both the number of samples and sequence length fixed. Specifically, we use 256 sequences of length 1024 sampled from RefinedWeb, Alpaca, or WikiText. The results are reported in Table~\ref{tab:calibration_domain}.

\begin{table*}[h]
\resizebox{\textwidth}{!}{%
\renewcommand{\arraystretch}{1.05}
\begin{tabular}{ccccccccccc:cc}
\hline
                                      &                                & \multicolumn{9}{c}{\textbf{Accuracy$\uparrow$}}                                                                                         & \multicolumn{2}{c}{\textbf{Perplexity$\downarrow$}} \\ \cline{3-13}
\multirow{-2}{*}{Calibration data}    & \multirow{-2}{*}{Method}       & \textbf{PIQA} & \textbf{HellaSwag} & \textbf{LAMBADA} & \textbf{ARC-e} & \textbf{ARC-c} & \textbf{SciQ} & \textbf{RACE} & \textbf{MMLU} & \textbf{Avg.} & \textbf{WikiText} & \textbf{LAMBADA} \\ \hline

                                      & SVD-LLM                        & 62.13 & 36.43 & 24.43 & 36.03 & 25.09 & 64.90 & 29.00 & 22.98 & 37.62 & 1.69E+02 & 1.70E+02 \\
\multirow{-2}{*}{RefinedWeb}          & CoSpaDi                         & \textbf{66.05} & \textbf{42.88} & \textbf{38.39} & \textbf{39.94} & \textbf{26.02} & \textbf{71.60} & \textbf{31.67} & \textbf{24.83} & \textbf{42.67} & \textbf{6.37E+01} & \textbf{3.51E+01} \\ \hdashline

                                      & SVD-LLM                        & 63.50 & 36.20 & 12.70 & 46.20 & 25.80 & 67.50 & 26.50 & 24.00 & 37.80 & 2.82E+02 & 9.24E+02 \\
\multirow{-2}{*}{Alpaca}              & CoSpaDi                         & \textbf{65.50} & \textbf{40.10} & \textbf{21.20} & \textbf{48.10} & \textbf{27.60} & \textbf{74.70} & \textbf{29.60} & \textbf{25.40} & \textbf{41.53} & \textbf{1.79E+02} & \textbf{2.60E+02} \\ \hdashline

                                      & SVD-LLM                        & 58.30 & 35.50 & 23.10 & 40.60 & 25.30 & 69.50 & 24.80 & \textbf{23.10} & 37.53 & 1.04E+02 & 1.91E+02 \\
\multirow{-2}{*}{WikiText}            & CoSpaDi                         & \textbf{59.50} & \textbf{38.10} & \textbf{31.20} & \textbf{41.80} & \textbf{26.60} & \textbf{73.40} & \textbf{27.60} & 23.00 & \textbf{40.15} & \textbf{5.89E+01} & \textbf{7.33E+01} \\ \hline
\end{tabular}
}
\caption{Sensitivity to the calibration domain on LLaMA-3.2-1B at $\mathrm{CR}=0.2$. All calibration sets contain 256 sequences of length 1024. We report accuracy on different benchmarks as well as its average and perplexity on WikiText and LAMBADA. Best results are provided in \textbf{bold}.}
\label{tab:calibration_domain}
\end{table*}

CoSpaDi consistently improves over SVD-LLM across all three calibration domains. The gains in average accuracy are $+5.05$, $+3.73$, and $+2.62$ percentage points for RefinedWeb, Alpaca, and WikiText, respectively. CoSpaDi also yields lower perplexity than SVD-LLM on both WikiText and LAMBADA for every calibration source considered. The absolute performance of both methods varies with the calibration domain, indicating that the choice of calibration data affects the resulting compressed model. Nevertheless, the relative advantage of CoSpaDi is preserved across all three domains.

\paragraph{Calibration set size.}
We next vary the amount of calibration data while fixing the calibration source to RefinedWeb and the sequence length to 1024. We evaluate 64, 128, and 256 calibration sequences. The results are shown in Table~\ref{tab:calibration_size}.

\begin{table*}[h]
\resizebox{\textwidth}{!}{%
\renewcommand{\arraystretch}{1.05}
\begin{tabular}{ccccccccccc:cc}
\hline
                                      &                                & \multicolumn{9}{c}{\textbf{Accuracy$\uparrow$}}                                                                                         & \multicolumn{2}{c}{\textbf{Perplexity$\downarrow$}} \\ \cline{3-13}
\multirow{-2}{*}{\# Calib. samples}   & \multirow{-2}{*}{Method}       & \textbf{PIQA} & \textbf{HellaSwag} & \textbf{LAMBADA} & \textbf{ARC-e} & \textbf{ARC-c} & \textbf{SciQ} & \textbf{RACE} & \textbf{MMLU} & \textbf{Avg.} & \textbf{WikiText} & \textbf{LAMBADA} \\ \hline

                                      & SVD-LLM                        & 59.30 & 35.30 & 19.80 & 35.10 & 24.10 & 58.80 & 28.30 & 23.10 & 35.48 & 2.03E+02 & 3.04E+02 \\
\multirow{-2}{*}{64}                  & CoSpaDi                         & \textbf{62.00} & \textbf{39.20} & \textbf{33.90} & \textbf{40.10} & \textbf{25.30} & \textbf{71.10} & \textbf{30.30} & \textbf{23.30} & \textbf{40.65} & \textbf{1.15E+02} & \textbf{6.38E+01} \\ \hdashline

                                      & SVD-LLM                        & 60.00 & 35.80 & 19.00 & 35.30 & 24.30 & 60.20 & 27.10 & 23.00 & 35.59 & 1.95E+02 & 3.16E+02 \\
\multirow{-2}{*}{128}                 & CoSpaDi                         & \textbf{62.80} & \textbf{39.80} & \textbf{31.20} & \textbf{39.00} & \textbf{25.20} & \textbf{68.30} & \textbf{30.40} & \textbf{23.20} & \textbf{39.99} & \textbf{1.15E+02} & \textbf{7.84E+01} \\ \hdashline

                                      & SVD-LLM                        & 62.13 & 36.43 & 24.43 & 36.03 & 25.09 & 64.90 & 29.00 & 22.98 & 37.62 & 1.69E+02 & 1.70E+02 \\
\multirow{-2}{*}{256}                 & CoSpaDi                         & \textbf{66.05} & \textbf{42.88} & \textbf{38.39} & \textbf{39.94} & \textbf{26.02} & \textbf{71.60} & \textbf{31.67} & \textbf{24.83} & \textbf{42.67} & \textbf{6.37E+01} & \textbf{3.51E+01} \\ \hline
\end{tabular}
}
\caption{Sensitivity to the number of calibration samples on LLaMA-3.2-1B at $\mathrm{CR}=0.2$ using RefinedWeb calibration data. All calibration sequences have length 1024. We report accuracy on different benchmarks as well as its average and perplexity on WikiText and LAMBADA. Best results are provided in \textbf{bold}.}
\label{tab:calibration_size}
\end{table*}

CoSpaDi consistently outperforms SVD-LLM at every calibration-set size considered, improving average accuracy by $+5.17$, $+4.40$, and $+5.05$ percentage points for 64, 128, and 256 calibration sequences, respectively. The advantage is therefore already present with only 64 calibration sequences and does not depend on using the largest calibration set considered in our main experiments. Increasing the amount of calibration data generally improves both methods, although the dependence is not strictly monotonic for every individual metric.

\paragraph{Summary.}
These experiments indicate that CoSpaDi's improvement over SVD-LLM is robust across both the calibration sources and calibration-set sizes considered here. Different calibration domains affect the absolute quality of the compressed models, but CoSpaDi consistently improves the average accuracy and perplexity relative to SVD-LLM. Similarly, substantial gains remain when reducing the calibration set from 256 to 64 sequences, suggesting that the observed advantage does not require a large calibration set.
\subsection{Evaluation Uncertainty}
\label{apx:evaluation_uncertainty}

The accuracy values reported throughout the paper are estimates obtained on finite benchmark test sets. To quantify the corresponding evaluation uncertainty, we use the task-level standard errors reported by \texttt{lm-evaluation-harness}. We consider LLaMA-3.2-1B at $\mathrm{CR}\in\{0.2,0.3,0.4\}$ under the same calibration data, compressed modules, and evaluation protocol as in the main experiments. For both SVD-LLM and CoSpaDi, the mean accuracies and corresponding standard errors are taken directly from our \texttt{lm-evaluation-harness} evaluation outputs.

Since \texttt{lm-evaluation-harness} reports a standard error separately for each benchmark, we estimate the standard error of the macro-average accuracy as
\begin{equation}
    \mathrm{SE}_{\mathrm{avg}}
    =
    \frac{1}{M}
    \sqrt{
        \sum_{i=1}^{M}
        \mathrm{SE}_i^2
    },
    \label{eq:average_accuracy_se}
\end{equation}
where $M=8$ is the number of accuracy benchmarks and $\mathrm{SE}_i$ is the standard error reported for benchmark $i$. This estimate treats the task-level evaluation uncertainties as independent.

For the difference in average accuracy between CoSpaDi and SVD-LLM, we combine the two method-level standard errors in quadrature,
using the same independence approximation:
\begin{equation}
    \mathrm{SE}_{\Delta}
    =
    \sqrt{
        \mathrm{SE}_{\mathrm{avg,CoSpaDi}}^2
        +
        \mathrm{SE}_{\mathrm{avg,SVD\text{-}LLM}}^2
    }.
    \label{eq:accuracy_gap_se}
\end{equation}

Table~\ref{tab:average_accuracy_uncertainty} compares the observed average-accuracy improvements with the corresponding combined standard errors.

\begin{table}[h]
\centering
\resizebox{0.45\textwidth}{!}{%
\renewcommand{\arraystretch}{1.05}
\begin{tabular}{ccccc}
\hline
\textbf{CR}
& \textbf{SVD-LLM Avg.}
& \textbf{CoSpaDi Avg.}
& \textbf{Gain}
& \textbf{Gain / $\mathrm{SE}_{\Delta}$} \\ \hline

0.2 & 0.3762 & 0.4267 & +0.0505 & $9.6\times$ \\
0.3 & 0.3022 & 0.3367 & +0.0346 & $6.6\times$ \\
0.4 & 0.2621 & 0.2753 & +0.0132 & $2.6\times$ \\ \hline

\end{tabular}
}
\caption{Average-accuracy gains relative to the combined standard error estimated from task-level \texttt{lm-evaluation-harness} outputs on LLaMA-3.2-1B. The displayed average accuracies are rounded to four decimal places;
gains are computed from the underlying unrounded averages.}
\label{tab:average_accuracy_uncertainty}
\end{table}

The CoSpaDi improvement is larger than the estimated evaluation uncertainty at all three compression ratios. Specifically, the average-accuracy gain corresponds to approximately $9.6\times$, $6.6\times$, and $2.6\times$ the combined standard error at $\mathrm{CR}=0.2$, $0.3$, and $0.4$, respectively. The margins are therefore particularly pronounced at $\mathrm{CR}=0.2$ and $0.3$, while the smaller improvement at $\mathrm{CR}=0.4$ also remains larger than the estimated benchmark-level uncertainty.

These estimates quantify uncertainty due to the finite number of evaluation examples and should not be interpreted as measuring the full run-to-run variability of the compression procedure. In particular, additional independent compression runs would be required to quantify variability arising from calibration-set sampling, dictionary initialization, and other sources of compression randomness.

\subsection{Number of Alternating Minimization Steps and Power Iteration Analysis}\label{apx:iter_ablation}

\begin{figure*}[!t]
    \centering
    \begin{subfigure}[t]{0.5\textwidth}
        \centering
        \includegraphics[width=\textwidth]{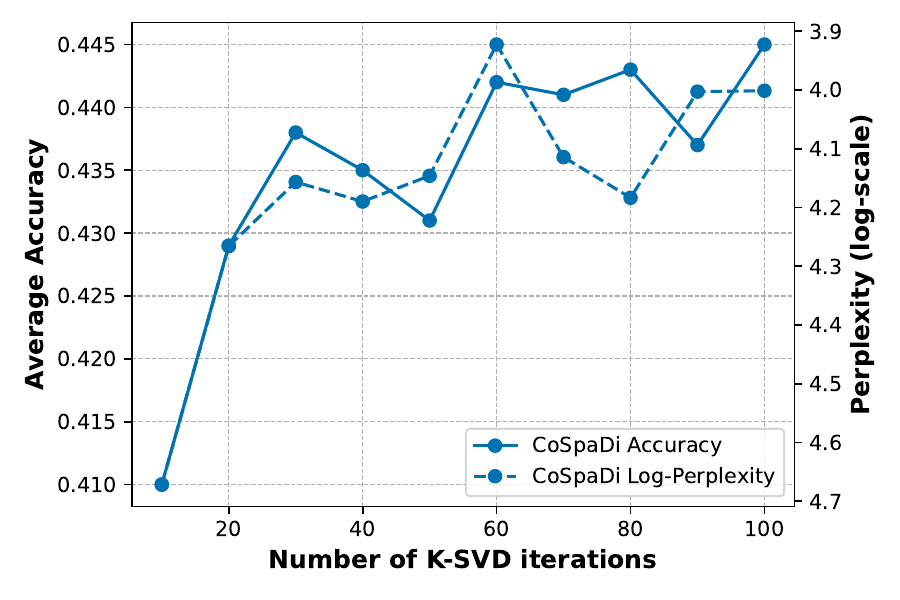}
        \label{fig:accuracy_perplexity_llama}
    \end{subfigure}%
    \hfill
    \begin{subfigure}[t]{0.5\textwidth}
        \centering
        \includegraphics[width=\textwidth]{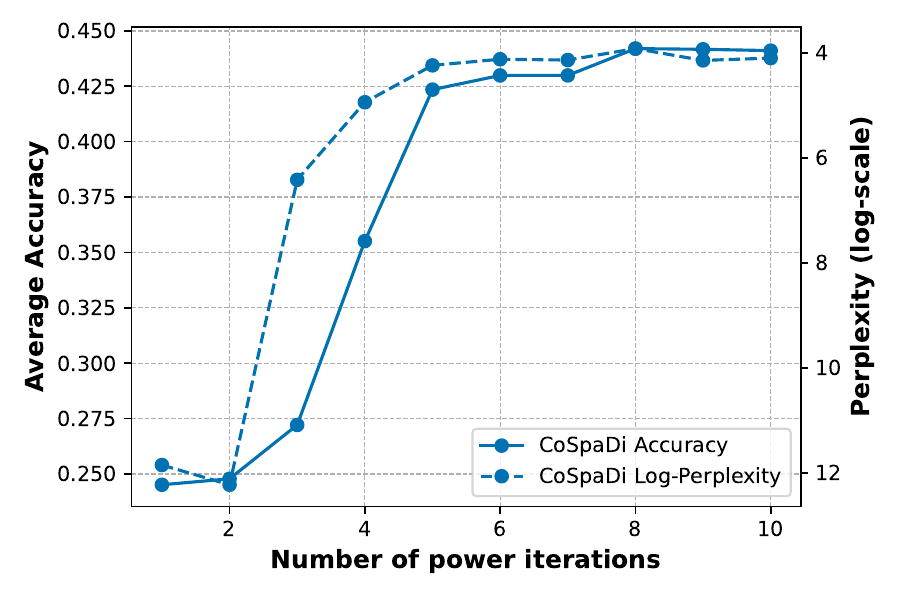}
    \end{subfigure}%
    \caption{Average benchmark accuracy and WikiText perplexity with respect to the number of K-SVD iterations (left) and the number of power iterations (right) for LLaMA-3.2-1B with $\rho=2$}
    \label{fig:KSVD_iters}
\end{figure*}

We conducted ablation studies to assess the effect of the number of K-SVD iterations and power iterations on performance using LLaMA-3.2-1B with fixed $\rho=2$. The left plot in Figure~\ref{fig:KSVD_iters} shows that most of the improvement is obtained within the first 40-50 K-SVD
iterations, after which performance fluctuates within a comparatively
narrow range. The right plot of Figure~\ref{fig:KSVD_iters} indicates that performance improves rapidly through approximately 5-6 power
iterations, with smaller changes thereafter.
Based on these results, we fixed the number of K-SVD iterations to 60 and power iterations to 8 in our final experiments, which provides a good balance between accuracy, perplexity, and computational efficiency.

\subsection{Runtime Breakdown and Scalability Discussion}\label{apx:runtime_breakdown}

The compression runtime of CoSpaDi is driven by the alternating optimization procedure, which combines OMP-based sparse coding with K-SVD-style dictionary updates.
To clarify the contribution of the main components, Table~\ref{tab:runtime_breakdown} reports a runtime breakdown for end-to-end compression of all targeted projection matrices in a 1B-scale model under the current solver configuration.

\begin{table}[t]
\centering
\resizebox{0.47\textwidth}{!}{%
\renewcommand{\arraystretch}{1.05}
\begin{tabular}{lcc}
\hline
Component & Time (min) & Share (\%) \\
\hline
Sparse coding (OMP) & 178.5 & 27.6 \\
Dictionary update (K-SVD) & 467.6 & 72.3 \\
Misc.\ overhead & 0.7 & 0.1 \\
\hline
Total & 646.8 & 100.0 \\
\hline
\end{tabular}
}
\caption{Runtime breakdown of CoSpaDi on a 1B-scale model for end-to-end compression of all targeted projection matrices under the current solver configuration.}
\label{tab:runtime_breakdown}
\end{table}

The results show that the dominant cost comes from the dictionary-update stage, while sparse coding accounts for a smaller but still substantial fraction of the overall runtime.
This behavior is consistent with the optimization structure of CoSpaDi, since K-SVD updates atoms sequentially through repeated rank-1 approximation subproblems.

Several acceleration directions are compatible with the current formulation.
First, online or stochastic dictionary learning can replace full-batch updates with incremental ones, improving scalability to larger problems~\citep{mairal2009online}.
Second, implementation-level optimizations such as Batch-OMP and efficient update procedures can substantially reduce wall-clock time in practice~\citep{rubinstein2008efficient}.
Third, the rank-1 update step can be approximated using truncated or randomized low-rank routines~\citep{halko2011randomized}, often combined with a small number of power iterations~\citep{golub2013matrix}, yielding improved accuracy--runtime trade-offs.

These observations suggest that the current compression time is primarily a property of the present solver implementation rather than of the CoSpaDi objective itself.
In this work, the focus is on establishing activation-aware sparse dictionary learning as a post-training compression paradigm and evaluating its compression--quality trade-offs; improving compression-time efficiency remains an important direction for future work.
\subsection{Detailed storage breakdown}\label{apx:storage_breakdown}

To make the storage accounting more transparent, we provide in Table~\ref{tab:storage_breakdown} a component-wise memory breakdown for CoSpaDi on the main settings.
Specifically, we separate the contribution of the dictionary, stored coefficient values, and mask/index overhead.
This complements the derivation in Appendix~\ref{apx:cr} and clarifies the relation between the full accounting of Eq.~\eqref{eq:cr_sd_mask} and the simplified accounting convention of Eq.~\eqref{eq:cr_sd_effective}.

The columns ``Applied eq.'' and ``Quant. coeff.'' indicate whether the target compression ratio is computed using the full accounting of Eq.~\ref{eq:cr_sd_mask} or the simplified convention of Eq.~\ref{eq:cr_sd_effective}, and whether coefficient truncation is applied.
Importantly, the table reports the actual component-wise allocation of storage in each case.
Thus, while the relative allocation between dictionary and coefficients changes across the two conventions, the total compressed footprint remains essentially unchanged up to rounding.

\begin{table*}[t]
\centering
\resizebox{0.7\textwidth}{!}{%
\renewcommand{\arraystretch}{1.05}
\begin{tabular}{llccccc}
\hline
\textbf{Model / Setting} & \textbf{Applied eq.} & \textbf{Quant. coeff.} & \textbf{Dictionary} & \textbf{Coeff. values} & \textbf{Mask / indices} & \textbf{Full total} \\
\hline
\textbf{LLaMA-3.2-1B} & -- & No  & --   & --   & --  & 1856 \\ \hline
\multirow{2}{*}{CoSpaDi, CR=20\%} & Eq.~10 & No  & 990  & 438  & 54  & 1484 \\
                 & Eq.~11 & Yes & 1018 & 408  & 58  & 1484 \\ \hdashline
\multirow{2}{*}{CoSpaDi, CR=30\%} & Eq.~10 & No  & 866  & 383  & 47  & 1298 \\
                 & Eq.~11 & Yes & 890  & 357  & 51  & 1298 \\ \hdashline
\multirow{2}{*}{CoSpaDi, CR=40\%} & Eq.~10 & No  & 743  & 329  & 41  & 1113 \\
                 & Eq.~11 & Yes & 763  & 306  & 43  & 1113 \\
\hline
\textbf{Qwen3-8B} & -- & No  & --   & --   & --  & 13248 \\ \hline
\multirow{2}{*}{CoSpaDi, CR=20\%} & Eq.~10 & No  & 7005 & 3192 & 399 & 10596 \\
                 & Eq.~11 & Yes & 7226 & 2948 & 421 & 10596 \\ \hdashline
\multirow{2}{*}{CoSpaDi, CR=30\%} & Eq.~10 & No  & 6129 & 2793 & 349 & 9272 \\
                 & Eq.~11 & Yes & 6322 & 2579 & 368 & 9271 \\ \hdashline
\multirow{2}{*}{CoSpaDi, CR=40\%} & Eq.~10 & No  & 5253 & 2393 & 299 & 7946 \\
                 & Eq.~11 & Yes & 5419 & 2211 & 315 & 7947 \\
\hline
\end{tabular}
}
\caption{Memory footprint breakdown (MiB) for storing all compressed linear layers (excluding embeddings, \texttt{lm\_head}, and normalization layers) under CoSpaDi at different compression ratios.}
\label{tab:storage_breakdown}
\end{table*}

For additional intuition, consider a single \texttt{up\_proj} matrix in Qwen3-8B with dimensions $d_1 = 4096$ and $d_2 = 12288$.

Under Eq.~10, the dictionary has shape $(4096 \times 3657)$, requiring approximately $28.57$ MiB.
The coefficient matrix has shape $(3657 \times 12288)$ with $1828$ nonzero entries per column, yielding approximately $42.84$ MiB for the stored coefficient values and $5.36$ MiB for the packed binary mask.

Under Eq.~11, the dictionary has shape $(4096 \times 3932)$, requiring approximately $30.72$ MiB.
The coefficient matrix has shape $(3932 \times 12288)$ with $1966$ nonzero entries per column, yielding approximately $40.32$ MiB for the truncated coefficient values and $5.76$ MiB for the packed binary mask.

This breakdown makes explicit how much each component contributes to the compressed representation and clarifies that the difference between the two accounting conventions lies primarily in how the storage budget is allocated between dictionary capacity and coefficient precision, rather than in the overall total size.
\subsection{Results for LLaMA-3.2-1B, Qwen3-0.6B and Qwen3-14B}\label{apx:more_models}

\begin{figure*}
\centering
\includegraphics[width=0.97\textwidth]{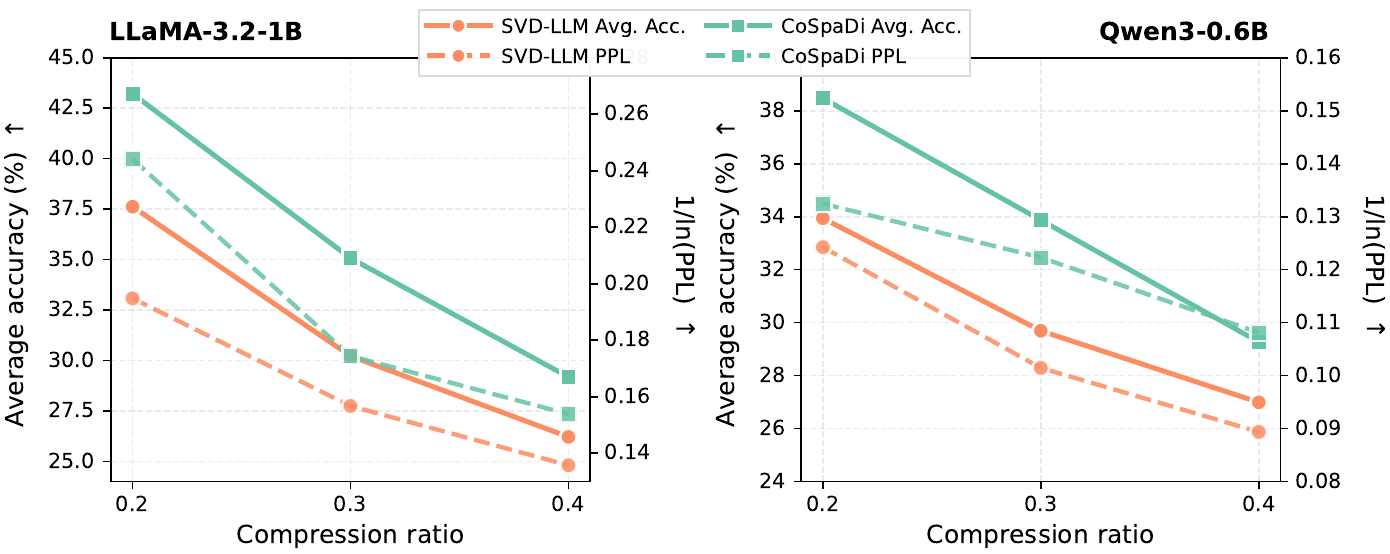}
\captionof{figure}{Average benchmark accuracy and WikiText perplexity for LLaMA-3.2-1B and Qwen3-0.6B using SVD-LLM and CoSpaDi with respect to compression ratio.}
\label{fig:small_models}
\end{figure*}

Figure~\ref{fig:small_models} summarizes the accuracy--compression and perplexity--compression curves for small models (LLaMA-3.2-1B and Qwen3-0.6B).

\begin{table*}[h]
\resizebox{\textwidth}{!}{%
\renewcommand{\arraystretch}{1.05}
\begin{tabular}{ccccccccccc:cc}
\hline
                            &                         & \multicolumn{9}{c}{\textbf{Accuracy$\uparrow$}}                                                                           & \multicolumn{2}{c}{\textbf{Perplexity$\downarrow$}}       \\ \cline{3-13}
\multirow{-2}{*}{\textbf{Method}}    & \multirow{-2}{*}{\textbf{CR}}    & \textbf{PIQA} & \textbf{HellaSwag} & \textbf{LAMBADA} & \textbf{ARC-e} & \textbf{ARC-c} & \textbf{SciQ} & \textbf{RACE}& \textbf{MMLU} & \textbf{Avg.} & \textbf{WikiText} & \textbf{LAMBADA} \\ \hline

\textbf{Qwen3-14B}          & --                     & 79.9 & 78.9 & 67.9 & 82.8 & 60.2 & 96.5 & 43.3 & 77.2 & 73.3 & 1.1E+01 & 3.7E+00 \\ \hline \hline

SVD-LLM                    &                         & 76.2 & 67.6 & 69.1 & 69.8 & 46.8 & 91.0 & 42.9 & 62.5 & 65.8 & 1.8E+01 & 4.1E+00 \\
CoSpaDi                    & \multirow{-2}{*}{0.2}   & 77.3 & 71.7 & 72.0 & 73.3 & 51.3 & 91.9 & 43.1 & 65.8 & \textbf{68.3} & \textbf{1.6E+01} & \textbf{3.4E+00} \\ \hline

SVD-LLM                    &                         & 72.0 & 59.7 & 61.5 & 62.6 & 39.3 & 88.9 & 41.5 & 52.0 & 59.7 & 2.3E+01 & 6.5E+00 \\
CoSpaDi                    & \multirow{-2}{*}{0.3}   & 74.6 & 64.2 & 69.0 & 65.3 & 43.6 & 90.7 & 42.5 & 58.0 & \textbf{63.5} & \textbf{2.0E+01} & \textbf{4.2E+00} \\ \hline

SVD-LLM                    &                         & 67.4 & 48.7 & 46.0 & 48.5 & 30.7 & 80.6 & 36.7 & 36.8 & 49.4 & 3.5E+01 & 1.8E+01 \\
CoSpaDi                    & \multirow{-2}{*}{0.4}   & 69.5 & 52.3 & 57.3 & 53.2 & 33.0 & 84.4 & 37.5 & 43.6 & \textbf{53.9} & \textbf{3.3E+01} & \textbf{8.7E+00} \\ \hline
\end{tabular}
}
\caption{Performance comparison of CoSpaDi vs SVD-LLM on Qwen3-14B at different compression levels on different benchmarks. Best results are provided in \textbf{bold}.}
\label{tab:main_llama3_8b}
\end{table*}

\begin{table*}[ht]
\resizebox{\textwidth}{!}{%
\renewcommand{\arraystretch}{1.05}
\centering
\begin{tabular}{ccccccccccc:cc}
\hline
                            &                         & \multicolumn{9}{c}{\textbf{Accuracy$\uparrow$}}                                                                           & \multicolumn{2}{c}{\textbf{Perplexity$\downarrow$}}       \\ \cline{3-13}
\multirow{-2}{*}{\textbf{Method}}    & \multirow{-2}{*}{\textbf{CR}}    & \textbf{PIQA} & \textbf{HellaSwag} & \textbf{LAMBADA} & \textbf{ARC-e} & \textbf{ARC-c} & \textbf{SciQ} & \textbf{RACE}& \textbf{MMLU} & \textbf{Avg.} & \textbf{WikiText} & \textbf{LAMBADA} \\ \hline

\textbf{LLaMA-3-8B} & --  & 80.7 & 79.1 & 75.6 & 77.7 & 53.5 & 93.9 & 40.3 & 62.2 & 70.4 & 7.3E+00 & 3.1E+00 \\ \hline \hline
SVD-LLM   & 0.4 & 60.3 & 34.5 & 11.4 & 32.4 & 24.5 & 44.2 & 25.7 & 23.1 & 32.0 & 5.5E+02 & 1.3E+03 \\
Wanda 2:4 & 0.44 & 62.4 & 39.7 & 0.8  & 41.9 & 25.6 & 74.7 & 26.8 & 23.4 & 36.9 & 2.1E+02 & 5.7E+03 \\
CoSpaDi   & 0.4 & 63.7 & 41.4 & 30.3 & 39.1 & 26.6 & 68.5 & 30.5 & 25.4 & \textbf{40.7} & \textbf{1.8E+02} & \textbf{1.2E+02} \\
\hline
\end{tabular}
}
\caption{Performance comparison on LLaMA-3-8B between SVD-LLM, CoSpaDi, and Wanda 2:4 at compression ratio 0.4. Best compressed result in each column is highlighted in \textbf{bold}.}
\label{tab:wanda_llama3_8b}
\end{table*}
\subsection{Additional Benchmark Results}
\label{apx:more_bench}
We additionally evaluate CoSpaDi and SVD-LLM on the Qwen model family using a set of more recent reasoning-oriented benchmarks.
Table~\ref{tab:more_bench} reports results for Qwen3-8B and Qwen3-14B across several compression ratios.
Overall, CoSpaDi matches or improves upon the SVD-LLM baseline on most benchmarks and settings, indicating that sparse dictionary learning provides a stronger compression--quality trade-off than low-rank factorization beyond the standard evaluation suite.

\begin{table}[ht]
\centering
\resizebox{0.47\textwidth}{!}{%
\renewcommand{\arraystretch}{1.05}
\begin{tabular}{c c c c c c c c}
\hline
\textbf{Model} & \textbf{CR} & \textbf{IFEval} & \textbf{BBH} & \textbf{MATH} & \textbf{GPQA} & \textbf{MUSR} & \textbf{MMLU-Pro} \\
      &     & (\%)   & (\%) & (\%) & (\%) & (\%) & (\%) \\
\hline
\textbf{Qwen3-8B} & ---      & 39.2 & 60.9 & 52.6 & 36.2 & 43.1 & 47.7 \\ \hline \hline
SVD-LLM  &                       & 25.5 & 41.0 & 1.1  & 28.4 & 39.8 & 26.3 \\
CoSpaDi & \multirow{-2}{*}{0.2} & \textbf{28.9} & \textbf{45.3} & \textbf{2.0}  & \textbf{28.6} & \textbf{42.1} & \textbf{31.5} \\ \hdashline
SVD-LLM  &                       & 22.9 & 34.4 & \textbf{1.0}  & \textbf{25.6} & \textbf{41.4} & 18.8 \\
CoSpaDi & \multirow{-2}{*}{0.3} & \textbf{25.2} & \textbf{38.2} & \textbf{1.0}  & 24.8 & 38.4 & \textbf{22.8} \\\hdashline
SVD-LLM  &                       & 22.7 & 30.2 & \textbf{0.8}  & 23.1 & 37.7 & 11.6 \\
CoSpaDi & \multirow{-2}{*}{0.4} & \textbf{26.1} & \textbf{32.7} & \textbf{0.8}  & \textbf{26.0} & \textbf{38.1} & \textbf{16.7} \\ \hline

\textbf{Qwen3-14B} & ---                & 43.3 & 63.2 & 53.3 & 38.6 & 40.7 & 53.3 \\ \hline \hline
SVD-LLM  &                       & 25.5 & 50.1 & 1.6  & 29.2 & \textbf{41.5} & 32.7 \\
CoSpaDi & \multirow{-2}{*}{0.2} & \textbf{29.6} & \textbf{54.2} & \textbf{4.2}  & \textbf{30.6} & 40.7 & \textbf{38.9} \\\hdashline
SVD-LLM  &                       & 23.0 & 34.3 & 1.1  & \textbf{26.4} & \textbf{39.7} & 12.8 \\
CoSpaDi & \multirow{-2}{*}{0.4} & \textbf{26.6} & \textbf{38.3} & \textbf{1.2}  & 25.5 & 38.0 & \textbf{15.7} \\ \hline

\end{tabular}
}
\caption{Performance comparison of CoSpaDi vs SVD-LLM on Qwen3-8B and Qwen3-14B at different compression levels on different benchmarks. Best results are provided in \textbf{bold}.}
\label{tab:more_bench}
\end{table}
\subsection{Comparison with a Semi-Structured Pruning Baseline}\label{apx:struct_pruning}

To further position CoSpaDi relative to pruning-based compression, we compare it with Wanda~\citep{sun2024wanda} in a 2:4 semi-structured setting on LLaMA-3-8B.
This baseline is particularly relevant from a deployment perspective because it follows a hardware-friendly N:M sparsity pattern.

For the 2:4 baseline, it is important to distinguish nominal sparsity from effective storage compression.
Although 2:4 sparsity removes 50\% of the weights, the realized memory reduction is smaller once metadata is taken into account.
In the standard semi-structured representation, each retained 16-bit weight requires an additional 2 bits of metadata, yielding an effective storage cost of 9 bits per original weight position and therefore an effective compression ratio of $1 - 9/16 \approx 43.8\%$, rather than the naive 50\%.
We therefore compare Wanda 2:4 to CoSpaDi at the closest matched storage budget.

Results are reported in Table~\ref{tab:wanda_llama3_8b}.
Overall, CoSpaDi remains competitive with pruning-based alternatives and provides the strongest average accuracy at the matched compression ratio.
In particular, CoSpaDi outperforms both SVD-LLM and Wanda 2:4 in average accuracy, while also achieving substantially better perplexity than both baselines on LAMBADA and better perplexity than SVD-LLM on WikiText.
More broadly, this comparison highlights that CoSpaDi should be viewed not only as an alternative to low-rank factorization, but also as a sparse reparameterization method that is naturally related to pruning.
At the same time, our current formulation is not limited to unstructured sparsity patterns: the same N:M constraints could in principle be imposed directly on the coefficient matrix, allowing each column to activate a hardware-friendly subset of dictionary atoms.
We view this as a promising direction for future work.

\begin{table*}[ht]
\centering
\resizebox{\textwidth}{!}{%
\renewcommand{\arraystretch}{1.05}
\begin{tabular}{lcccccccccc:cc}
\hline
\multirow{2}{*}{\textbf{Method}} 
& \multirow{2}{*}{\textbf{Global CR}} 
& \multicolumn{9}{c}{\textbf{Accuracy}$\uparrow$} 
& \multicolumn{2}{c}{\textbf{Perplexity}$\downarrow$} \\
\cline{3-13}
& & \textbf{PIQA} & \textbf{HellaSwag} & \textbf{LAMBADA} & \textbf{ARC-e} & \textbf{ARC-c} & \textbf{SciQ} & \textbf{Race} & \textbf{MMLU} & \textbf{Avg.} & \textbf{Wiki} & \textbf{LAMBADA} \\
\hline
\textbf{LLaMA-3-8B}              & --   & 80.7 & 79.1 & 75.6 & 77.7 & 53.5 & 93.9 & 40.3 & 62.2 & 70.4 & 7.3E+00 & 3.1E+00 \\ \hline \hline
CoSpaDi                 & 0.20 & 75.2 & 66.5 & 73.8 & 66.5 & 41.6 & 89.5 & 38.2 & 42.8 & 61.8 & 2.0E+01 & 4.3E+00 \\ \hdashline
CoSpaDi$^{\mathrm{UG}}$ & 0.22 & 75.3 & 63.2 & 76.7 & 67.8 & 42.1 & 92.5 & 37.3 & 40.2 & \textbf{61.9} & \textbf{2.5E+01} & \textbf{2.8E+00} \\
SVD-LLM$^{\mathrm{UG}}$ & 0.22 & 71.9 & 51.9 & 56.7 & 58.8 & 33.5 & 87.7 & 33.8 & 33.9 & 53.5 & 5.6E+01 & 8.5E+00 \\
\hline
CoSpaDi                 & 0.30 & 70.5 & 56.2 & 61.3 & 54.2 & 33.5 & 85.7 & 36.2 & 32.2 & 53.7 & \textbf{4.5E+01} & 9.2E+00 \\ \hdashline
CoSpaDi$^{\mathrm{UG}}$ & 0.28 & 70.1 & 51.5 & 64.3 & 58.4 & 35.3 & 89.7 & 34.5 & 34.7 & \textbf{54.8} & 5.2E+01 & \textbf{5.8E+00} \\
SVD-LLM$^{\mathrm{UG}}$ & 0.28 & 65.0 & 40.9 & 33.0 & 48.1 & 28.3 & 79.9 & 28.8 & 26.1 & 43.8 & 2.1E+02 & 5.5E+01 \\
\hline
\end{tabular}
}
\caption{Quality of selective \texttt{up}/\texttt{gate} compression on LLaMA-3-8B. Superscript $\mathrm{UG}$ denotes compressing only the MLP \texttt{up} and \texttt{gate} projections.}
\label{tab:deployment_quality}
\end{table*}

\subsection{From Theory to Practice: Deployment Considerations}
\label{apx:deployment}

While CoSpaDi improves the accuracy--compression trade-off in the uniform compression setting, practical deployment also depends on where compression is applied.
Not all projection layers contribute equally to parameter count, accuracy degradation, or runtime.
In modern LLMs, MLP expansion projections are substantially wider than attention projections, making the \texttt{up} and \texttt{gate} matrices especially attractive targets for selective compression.
We therefore evaluate a hardware-aware setting in which only these two projection types are replaced by compressed modules.
We denote such variants with the superscript $\mathrm{UG}$.

This selective setting is favorable for CoSpaDi because the wide \texttt{up}/\texttt{gate} matrices are naturally represented as a dense dictionary followed by sparse coefficients.
In contrast to uniform compression, selective compression concentrates the storage budget on the layers that offer the largest parameter savings and the most suitable dense-times-sparse computation pattern.
For CoSpaDi$^{\mathrm{UG}}$, we use pretrained compressed checkpoints at the corresponding layer-wise compression ratio and replace only the \texttt{up} and \texttt{gate} modules in the dense model.
We compare against SVD-LLM$^{\mathrm{UG}}$ under the same replacement protocol.

Our inference experiments use a lightweight PyTorch/HuggingFace
Transformers module-replacement setup. We did not develop a custom
production CUDA kernel and do not integrate CoSpaDi into vLLM or TGI.
The sparse coefficient computation uses the MACKO-backed path available
to us, which is primarily designed for sparse matrix--vector
multiplication. Consequently, the reported measurements should be
interpreted as a research deployment evaluation rather than an optimized
production-serving result; efficient batched serving would additionally
require sparse--dense matrix--matrix kernels and runtime integration.

\paragraph{Quality of selective compression.}
Table~\ref{tab:deployment_quality} reports the resulting accuracy and perplexity on LLaMA-3-8B.
Selective CoSpaDi preserves the quality of the uniformly compressed CoSpaDi model at a similar global compression ratio, and substantially outperforms SVD-LLM$^{\mathrm{UG}}$ at matched global CR.
This indicates that the union-of-subspaces representation remains beneficial even when compression is restricted to deployment-favorable projection types.

\paragraph{Synchronous throughput.}
We first measure synchronous generation throughput on an RTX 3090 across prompt lengths from 1 to 256 tokens.
Table~\ref{tab:deployment_sync_throughput} reports tokens/s for SVD-LLM$^{\mathrm{UG}}$ and CoSpaDi$^{\mathrm{UG}}$ at several \texttt{up}/\texttt{gate} compression ratios.
SVD-LLM$^{\mathrm{UG}}$ is faster in raw tokens/s under the current implementation, but CoSpaDi$^{\mathrm{UG}}$ remains close in throughput while providing substantially stronger model quality at matched global CR.

\begin{table}[t]
\centering
\resizebox{0.9\columnwidth}{!}{%
\renewcommand{\arraystretch}{1.05}
\begin{tabular}{cccc}
\hline
\textbf{Prompt} & \textbf{UG CR} & \textbf{SVD-LLM$^{\mathrm{UG}}$} & \textbf{CoSpaDi$^{\mathrm{UG}}$} \\
\hline
1   & \multirow{6}{*}{0.3}  & 51.2 & 46.8 \\
16  &  & 50.3 & 46.8 \\
32  &  & 51.3 & 45.9 \\
64  &  & 50.3 & 45.3 \\
128 &  & 50.0 & 44.8 \\
256 &  & 49.5 & 44.0 \\
\hline
\textbf{Avg.} &  & \textbf{50.4} & 45.6 \\
\hline
1   & \multirow{6}{*}{0.4} & 53.3 & 48.4 \\
16  &  & 53.5 & 48.3 \\
32  &  & 54.4 & 48.0 \\
64  &  & 54.2 & 48.4 \\
128 &  & 52.7 & 47.8 \\
256 &  & 51.9 & 47.3 \\
\hline
\textbf{Avg.} &  & \textbf{53.3} & 48.0 \\
\hline
1   & \multirow{6}{*}{0.5} & 53.8 & 53.0 \\
16  &  & 52.4 & 52.0 \\
32  &  & 52.9 & 49.1 \\
64  &  & 52.4 & 50.4 \\
128 &  & 52.5 & 49.5 \\
256 &  & 50.9 & 49.0 \\
\hline
\textbf{Avg.} &  & \textbf{52.5} & 50.5 \\
\hline
\end{tabular}
}
\caption{Synchronous generation throughput on LLaMA-3-8B using an RTX 3090. Throughput is reported in tokens/s; higher is better.}
\label{tab:deployment_sync_throughput}
\end{table}

\paragraph{Asynchronous throughput.}
We also evaluate asynchronous serving with concurrent users.
Table~\ref{tab:deployment_async_prompt1} compares the dense model, SVD-LLM$^{\mathrm{UG}}$, and CoSpaDi$^{\mathrm{UG}}$ for prompt length 1.
Both compressed variants improve throughput over the dense baseline for most concurrency settings, and CoSpaDi$^{\mathrm{UG}}$ achieves the highest throughput at the largest tested concurrency.

\begin{table}[ht]
\centering
\resizebox{\columnwidth}{!}{%
\renewcommand{\arraystretch}{1.05}
\begin{tabular}{lcccc}
\hline
\textbf{Method} & \textbf{1 user} & \textbf{4 users} & \textbf{8 users} & \textbf{16 users} \\
\hline
Dense baseline & 77.0 & 290.0 & 589.0 & 1344.0 \\
SVD-LLM$^{\mathrm{UG}}$ & \textbf{99.4} & \textbf{386.0} & \textbf{798.0} & 1332.0 \\
CoSpaDi$^{\mathrm{UG}}$ & 95.6 & 375.6 & 758.0 & \textbf{1515.0} \\
\hline
\end{tabular}
}
\caption{Asynchronous serving throughput on LLaMA-3-8B for prompt length $1$. SVD-LLM$^{\mathrm{UG}}$ and CoSpaDi$^{\mathrm{UG}}$ use UG CR $=0.60$ and $\rho=2$; higher is better.}
\label{tab:deployment_async_prompt1}
\end{table}

Table~\ref{tab:deployment_async_grid} reports the full asynchronous prompt-length sweep for CoSpaDi$^{\mathrm{UG}}$.
Throughput remains stable across prompt lengths and scales with the number of concurrent users, indicating that selective CoSpaDi compression is compatible with multi-request serving.

\begin{table}[ht]
\centering
\resizebox{\columnwidth}{!}{%
\renewcommand{\arraystretch}{1.05}
\begin{tabular}{cccccc}
\hline
\textbf{Users} & \textbf{Prompt 1} & \textbf{Prompt 16} & \textbf{Prompt 32} & \textbf{Prompt 64} & \textbf{Prompt 128} \\
\hline
1  & 95.6  & 93.5  & 92.4  & 101.0 & 98.0  \\
4  & 375.6 & 374.8 & 372.4 & 393.0 & 401.0 \\
8  & 758.0 & 742.0 & 749.4 & 802.0 & 792.0 \\
16 & 1515.0 & 1479.6 & 1472.0 & 1521.0 & 1573.0 \\
\hline
\end{tabular}
}
\caption{Asynchronous serving throughput of CoSpaDi$^{\mathrm{UG}}$ on LLaMA-3-8B with UG CR $=0.60$ and $\rho=2$. Throughput is reported in tokens/s; higher is better.}
\label{tab:deployment_async_grid}
\end{table}

Overall, selective \texttt{up}/\texttt{gate} compression shows that CoSpaDi is not only a stronger quality-preserving compression method, but also a practical deployment candidate.
Its current throughput is competitive with a low-rank counterpart while retaining much higher accuracy and lower perplexity at matched global CR.
Since CoSpaDi evaluates projections as $(\m X\m D_a)\m S$, future improvements in sparse matrix kernels, sparse coefficient layouts, and inference runtimes can directly improve its latency profile.

\end{document}